\documentclass[11pt]{article}

\usepackage[final]{acl}

\usepackage{times}
\usepackage{latexsym}
\usepackage[utf8]{inputenc}

\usepackage{makecell}
\usepackage{tabularx}

\usepackage[T1]{fontenc}
\usepackage[utf8]{inputenc}

\usepackage{microtype}

\usepackage{inconsolata}

\usepackage{graphicx}
\usepackage{xcolor}
\usepackage{amsmath, amssymb}
\usepackage{tcolorbox}
\tcbuselibrary{listings,skins,breakable}

\usepackage[dvipsnames,table,xcdraw]{xcolor}
\usepackage{lipsum}               % 用于生成示例文字
\tcbuselibrary{breakable}
\usepackage{cuted}
\usepackage{fancyvrb}
\usepackage{listings}
\usepackage{booktabs}
\usepackage{multirow}
\usepackage{geometry}
\usepackage{graphicx}
\usepackage{subcaption}
\usepackage{float}
\usepackage{rotating} 
\usepackage{caption}    
\usepackage{xcolor}        
\usepackage{tcolorbox}
\tcbuselibrary{breakable, skins}
\newtcblisting{tcbverbatim}{
    listing only,
    listing options={basicstyle=\ttfamily},
    colback=gray!10,
    colframe=gray!50,
    breakable
}

\newtcolorbox{examplebox}{
  enhanced,
  breakable,
  colback=gray!3,
  colframe=black,
  boxrule=0.8pt,
  arc=2pt,
  left=6pt,
  right=6pt,
  top=6pt,
  bottom=6pt
}

\title{SciExplore: Evaluating Autonomous Agents from Scientific Navigation \\ to Information Integration}

\author{
    \textbf{Yinhao Tang}$^{1,2}$\textsuperscript{*} \quad
    \textbf{Youqing Fang}$^{1,2}$\textsuperscript{*} \quad
    \textbf{Yanan Sun}$^{2}$\textsuperscript{\dag} \quad \\
    \textbf{Wenran Liu}$^{2}$\quad
    \textbf{Weiming Zhang}$^{1}$\quad 
    \textbf{Bin Liu}$^{1}$ \quad \\
    \textbf{Kuikun Liu}$^{2}$ \quad
    \textbf{Wenwei Zhang}$^{2}$ \quad
    \textbf{Kai Chen}$^{2}$\textsuperscript{\dag} \quad \\
    \textsuperscript{1}University of Science and Technology of China \textsuperscript{2}Shanghai AI Laboratory \\
    \normalsize
    \normalfont
    \texttt{\{tangyinhao,fangyq\}@mail.ustc.edu.cn}, 
    \texttt{\{sunyanan,chenkai\}@pjlab.org.cn}
}

\begin{document}
\maketitle

\begingroup
\renewcommand{\thefootnote}{}
\footnotetext{* Equal contribution.\, $\dag$ Corresponding author.}
\endgroup

\begin{figure*}[t]
  \centering
  \includegraphics[width=\textwidth]{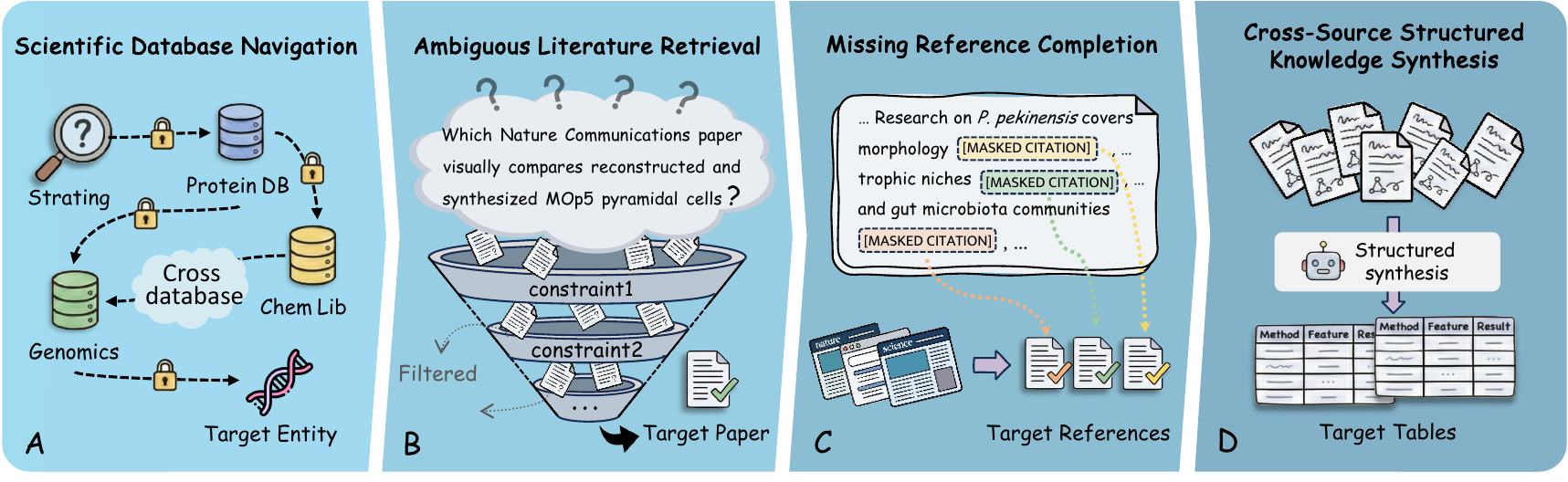}
  % \vspace{-1.2 em}
  \caption{\textbf{Overview of \textsc{SciExplore}.} 
    The benchmark evaluates scientific information seeking as a progressive cognitive process, spanning four task types that advance from (A) entity-level reasoning, to (B) document-level literature identification, (C) evidence-level reference grounding, and finally (D) domain-level knowledge synthesis.}
    \label{fig:teaser}
    % \vspace{-1.25em}
\end{figure*}

\begin{abstract}

Scientific research involves complex information-seeking and reasoning workflows across heterogeneous sources.
However, existing benchmarks primarily emphasize general-domain retrieval or static scientific question answering, and therefore fail to assess key capabilities required in realistic scientific research workflows.
We introduce SciExplore, a benchmark designed to evaluate scientific information-seeking and reasoning capabilities of LLMs and agents.
SciExplore comprises four task types covering 103 expert-curated tasks across more than ten scientific disciplines: scientific database navigation, ambiguous literature retrieval, missing reference completion, and cross-source structured knowledge synthesis, which probe progressively higher-level abilities from entity-level reasoning and document-level identification to evidence-level grounding and domain-level synthesis.
We evaluate over ten state-of-the-art LLMs and autonomous agents on SciExplore, revealing substantial performance gaps with performance degrading sharply as task complexity increases and extremely low accuracy on the most challenging structured synthesis tasks.
These results highlight significant limitations of current models and agents in realistic scientific information-seeking scenarios.
\end{abstract}
\section{Introduction}
As scientific discovery becomes increasingly data-intensive and interdisciplinary, Large Language Models (LLMs) are envisioned as autonomous scientific assistants~\cite{schmidgall2025agent, baek2025researchagent, chai2025scimaster, li2025websailor, wu2025webdancer, li2025webthinker, bran2023chemcrow, li2025reseek, huang2025manusearch} capable of supporting complex research workflows. These workflows extend well beyond fact lookup~\cite{yang2018hotpotqa, wei2025browsecomp, zhou2025browsecomp} or question answering~\cite{saikh2022scienceqa, wang2023scibench, yue2024mmmu}, encompassing tasks such as navigating specialized databases, identifying relevant literature under ambiguous descriptions, grounding claims in appropriate evidence, and synthesizing structured knowledge across heterogeneous sources. A central challenge for the community, therefore, is to rigorously evaluate whether current LLMs or agents can reliably perform these research behaviors in realistic settings.

Despite rapid progress, existing evaluation paradigms remain poorly aligned with the demands of authentic scientific research. On one hand, deep search benchmarks~\cite{wei2025browsecomp, zhou2025browsecomp, wu2025webwalker, wong2025widesearch} primarily target general-domain information retrieval and open-web exploration, overlooking the structured databases, domain-specific conventions, and methodological reasoning intrinsic to scientific inquiry. Deep Research benchmarks~\cite{du2025deepresearch} extend this paradigm by evaluating agents through long-form research report generation, with primary emphasis on the completeness and fluency of written reports rather than fine-grained evidence grounding or structured knowledge synthesis. On the other hand, science-oriented QA benchmarks~\cite{saikh2022scienceqa, wang2023scibench, yue2024mmmu, he2024cmmu, guo2023can, cui2025curie, du2025supergpqa, zhou2025scientists, wang2022scienceworld} typically reduce evaluation to static question answering in closed environments, abstracting away crucial steps such as evidence discovery, literature disambiguation, and cross-source integration. As a result, neither paradigm adequately measures the capabilities required of LLMs as research assistants—namely, evidence-driven, multi-step reasoning grounded in real, noisy, and heterogeneous scientific information environments.

% To bridge this gap, we introduce \textsc{SciExplore}, an expert-curated benchmark designed to evaluate LLMs or agents within authentic scientific information-seeking and reasoning workflows. The design of \textsc{SciExplore} is motivated by a key observation: scientific information seeking is inherently a progressive cognitive process. Researchers must incrementally advance from locating individual entities, to identifying relevant documents, grounding claims in supporting evidence, and ultimately synthesizing structured knowledge at the domain level. Capturing this progression is essential for evaluating whether agents can function as reliable scientific assistants rather than isolated retrieval tools.
To address this gap, we introduce \textsc{SciExplore}, an expert-curated benchmark for evaluating LLMs or agents in authentic scientific information-seeking and reasoning workflows. \textsc{SciExplore} is designed based on the observation that scientific information seeking is a progressive cognitive process where researchers incrementally move from locating individual entities to identifying relevant documents, grounding claims in evidence, and synthesizing structured knowledge at the domain level. % This progression is crucial for assessing whether agents can serve as reliable scientific assistants rather than mere retrieval tools.

Accordingly, \textsc{SciExplore} comprises four task types, each targeting a distinct and essential scientific capability, as illustrated in Fig.~\ref{fig:teaser}.
\textbf{(T1)} \textit{Scientific Database Navigation} evaluates the foundational ability to traverse structured scientific databases,
%~\cite{Kim2023_PubChem, Ochoa2023_OpenTargets, Jain2013, UniProt2025, Zdrazil2024_ChEMBL,Knox2024_DrugBank}, 
requiring multi-step reasoning to locate precise entities and attributes across interconnected records.
\textbf{(T2)} \textit{Ambiguous Literature Retrieval} builds on this foundation by shifting from databases to the scientific literature, challenging agents to identify a specific key paper based on vague, incomplete, or noisy methodological descriptions, without relying on explicit keywords.
\textbf{(T3)} \textit{Missing Reference Completion} further increases difficulty by moving from single-document identification to evidence-level grounding: given a scientific paragraph with missing citations, agents must correctly align multiple claims with their appropriate supporting references.
\textbf{(T4)} \textit{Cross-Source Structured Knowledge Synthesis} represents the highest level of abstraction, requiring agents to aggregate evidence from a broad set of papers and synthesize these findings into a structured comparison table that captures a domain-level overview of a research topic.

\begin{table*}[t]
\centering
\caption{Comparison of \textsc{SciExplore} with existing benchmarks.
Abbreviations:
\textbf{Sci. K. \& R.}: Scientific Knowledge and Reasoning;
\textbf{Sci. DB \& Lit.}: Scientific Database and Literature;
\textbf{ST}: Short Text;
\textbf{MC}: Multi Choice;
\textbf{Ref.}: Bibliographic Reference.}
\resizebox{\textwidth}{!}{%
\begin{tabular}{lcccccc}
\toprule
\textbf{Benchmark} &
\textbf{Domain} &
\textbf{Sci. K. \& R.} &
\textbf{Search Depth} &
\textbf{Search Width} &
\textbf{Search Source} &
\textbf{Output Format} \\
\midrule

\rowcolor{gray!15}
\multicolumn{7}{l}{\textit{Deep Search and Deep Research Benchmarks}} \\

HotpotQA~\cite{yang2018hotpotqa} & General & Low & Low & Low & Wikipedia & ST \\
BrowseComp~\cite{wei2025browsecomp} & General & Low & High & Low & Open Web & ST \\
WideSearch~\cite{wong2025widesearch} & General & Low & Low & High & Open Web & Table \\
DeepResearchBench~\cite{du2025deepresearch} & General & Low & Middle & Middle & Open Web & Report \\

\midrule
\rowcolor{gray!15}
\multicolumn{7}{l}{\textit{Scientific QA Benchmarks}} \\

ScienceQA~\cite{saikh2022scienceqa} & Science & Low & Low & Low & Open Web  & MC \\
SciBench~\cite{wang2023scibench} & Science & Middle & Middle & Low & Open Web  & ST \\
SuperGPQA~\cite{du2025supergpqa} & Science & High & High & Low & Open Web  & ST \\
HLE~\cite{phan2025humanity} & Math \& Science & High & High & Low & Open Web  & ST \\

\midrule
\rowcolor{gray!15}
\multicolumn{7}{l}{\textit{Our proposed}} \\

\textbf{\textsc{SciExplore}} & \textbf{Science} & \textbf{High} & \textbf{High} & \textbf{High} &
\textbf{Sci. DB \& Lit.} & \textbf{ST \& Ref. \& Table} \\

\bottomrule
\end{tabular}%
}
\label{tab:benchmark-comparison}
% \vspace{-1em}
\end{table*}

% Together, these tasks form a closed-loop evaluation of scientific information seeking, progressing systematically from entity-level retrieval, to document-level identification, to evidence-level grounding, and finally to domain-level knowledge synthesis. Based on this design, we construct 103 PhD-level tasks curated by domain experts across more than ten scientific disciplines. \textsc{SciExplore} enforces strict difficulty control and answer uniqueness, ensuring that successful completion requires genuine domain understanding and multi-step reasoning rather than reliance on parametric memorization or shortcut retrieval.

% We evaluate 12 state-of-the-art LLMs and autonomous search agents on \textsc{SciExplore}. The results reveal pronounced capability gaps: most systems achieve overall success rates below 50\%, with performance degrading sharply as task complexity increases. Notably, on the most challenging task—Cross-Source Structured Knowledge Synthesis—even the strongest agents exhibit near-zero table-level accuracy. These findings indicate that while current systems can handle isolated retrieval or shallow reasoning, they struggle substantially with realistic scientific research tasks that demand broad evidence coverage, precise grounding, and high-fidelity structured synthesis.

This study outlines a closed-loop evaluation of scientific information seeking, moving systematically from entity retrieval to knowledge synthesis, involving 103 PhD-level tasks curated by experts across various disciplines. The platform, \textsc{SciExplore}, maintains strict difficulty levels and ensures answer uniqueness, requiring deep domain understanding and multi-step reasoning instead of simple memorization.

We assessed 12 advanced LLMs and search agents on \textsc{SciExplore}, revealing significant performance gaps: most achieved success rates under 50\%, with a sharp decline in effectiveness as task complexity increased. Notably, the toughest task—Cross-Source Structured Knowledge Synthesis—showed even top agents with accuracy lower than 20\%. These results indicate that while current systems manage basic retrieval and shallow reasoning, they significantly falter in complex scientific research that requires comprehensive evidence, precise grounding, and accurate synthesis.

In summary, our contributions are threefold:
\begin{itemize}
\item We introduce \textsc{SciExplore}, a novel benchmark that evaluates agents along a hierarchy of scientific reasoning capabilities, spanning entity-level database navigation, document-level literature identification, evidence-level reference grounding, and domain-level structured knowledge synthesis.
\item We propose a rigorous, expert-driven data construction and validation framework that enforces high reasoning depth, strong domain specificity, and answer uniqueness, effectively preventing shortcut-based or memorization-driven solutions.
\item Through extensive evaluation of state-of-the-art agents, we expose substantial limitations in current systems when deployed in realistic scientific research scenarios, particularly their inability to reliably integrate multi-source evidence into precise, structured outputs.
\end{itemize}
\section{Related Works}

\begin{figure*}[t]
    \centering
    \includegraphics[width=\textwidth]{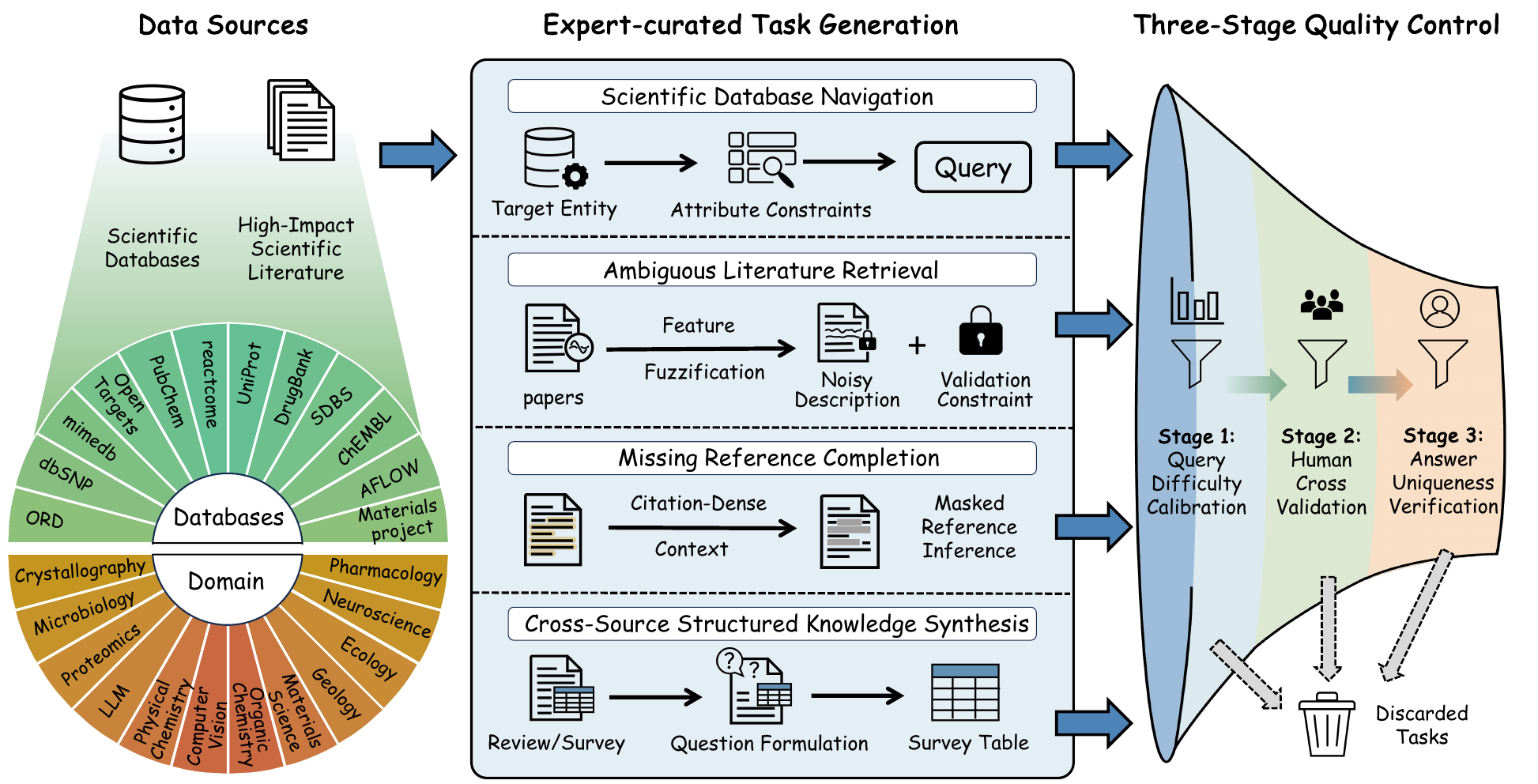}
    \caption{\textbf{Dataset construction pipeline of \textsc{SciExplore}.}
    The pipeline illustrates expert-driven task design, multi-stage validation, and rigorous quality control across T1–T4.}
    \label{fig:dataset-pipeline}
    % \vspace{-1em}
\end{figure*}

\noindent\textbf{Deep Search and Deep Research Benchmarks.}
Recent benchmarks for evaluating search agents in the general domain have progressively moved beyond single-hop fact retrieval toward deeper multi-step search, planning, and synthesis.
Early QA datasets~\cite{kwiatkowski2019natural,yang2018hotpotqa} focus on limited multi-hop reasoning, while later benchmarks explicitly introduce tool use and long-horizon search behaviors~\cite{mialon2023gaia, chen2025xbench}.
More recent work expands the evaluation scope along complementary dimensions.
BrowseComp-EN/ZH~\cite{wei2025browsecomp, zhou2025browsecomp} emphasize search depth and robustness to noise and entity ambiguity, whereas WideSearch~\cite{wong2025widesearch} stresses search breadth and structured aggregation from large-scale web sources.
DeepResearchBench~\cite{du2025deepresearch} further pushes this paradigm by requiring agents to synthesize multi-source evidence into long-form research-style reports.

Despite these advances, existing deep search benchmarks remain largely confined to the general web domain, relying on open-web sources and common-sense fact aggregation, and thus failing to evaluate core scientific assistant capabilities such as database navigation, interpretation of ambiguous methodological contexts, and evidence-level grounding.
Moreover, synthesis is typically assessed through free-form text, avoiding the stricter requirement of producing precise, structured scientific artifacts with high-fidelity evidence alignment (Table~\ref{tab:benchmark-comparison}), leaving a gap in evaluating agents in authentic scientific information-seeking workflows.

\noindent{\textbf{Scientific QA Benchmarks.}}
Scientific QA benchmarks primarily evaluate models’ parametric scientific reasoning under closed-book or minimally grounded settings.
Early benchmarks target high-school or undergraduate-level science questions~\cite{saikh2022scienceqa, hendrycks2020measuring}, while more recent datasets such as SciBench~\cite{wang2023scibench} introduce college-level problems involving multi-step calculations.
Advanced benchmarks including GPQA~\cite{rein2024gpqa}, SuperGPQA~\cite{du2025supergpqa}, and HLE~\cite{phan2025humanity} further push difficulty toward expert-level and frontier scientific reasoning.

While effective at measuring reasoning depth, existing scientific QA benchmarks do not explicitly evaluate information seeking, evidence discovery, or cross-source validation, as high performance can often be achieved using parametric knowledge alone.
In contrast, \textsc{SciExplore} is constructed to favor problems whose resolution requires substantial retrieval and filtering from external scientific sources, encouraging agents to acquire missing knowledge and ground their reasoning in externally obtained evidence for structured scientific~synthesis.

% \noindent{\textbf{Positioning of \textsc{SciExplore}.}} Taken together, prior benchmarks tend to emphasize either search competence in general-domain settings or scientific reasoning without explicit information seeking. \textsc{SciExplore} is designed to sit at the intersection of these two paradigms by evaluating progressive scientific information-seeking capabilities, ranging from entity-level database navigation and document-level literature identification to evidence-level grounding and domain-level structured knowledge synthesis. This positioning supports a more faithful assessment of LLM-based agents as scientific assistants operating in realistic research workflows.
\section{\textsc{SciExplore}}
% To operationalize the desiderata above, \textsc{SciExplore} instantiates scientific information-seeking as four complementary tasks. We discuss the task definitions in Section~\ref{sec:task_definition}, data construction in Section~\ref{sec:dataset_construction}, and quality control in Section~\ref{sec:quality_control}. Detailed data statistics and evaluation metrics can be found in Appendix~\ref{appendix_dataset_distribution} and Appendix~\ref{app:evaluation_details}, respectively.

To operationalize the desiderata above, \textsc{SciExplore} instantiates scientific information-seeking as four complementary tasks, curated through a fully expert-driven construction and verification pipeline. We formalize this task taxonomy in Section~\ref{sec:task_definition}, detail the corresponding data construction procedure in Section~\ref{sec:dataset_construction}, and describe our quality control mechanisms in Section~\ref{sec:quality_control}. Detailed data statistics and evaluation metrics are further provided in Appendix~\ref{appendix_dataset_distribution} and Appendix~\ref{app:evaluation_details}, respectively.

% To operationalize the desiderata above, \textsc{SciExplore} instantiates scientific information-seeking as four complementary tasks---spanning database navigation, ambiguous literature retrieval, missing reference completion, and cross-source structured knowledge synthesis---that jointly stress an agent's ability to locate, disambiguate, and integrate evidence from authentic scientific sources. Every instance is authored and verified by domain experts, and the full pipeline is designed to enforce realism, answer uniqueness, and resistance to parametric shortcuts, so that strong performance necessarily reflects genuine retrieval and reasoning rather than memorization. 
% We discuss the task definitions in Section~\ref{sec:task_definition}, data construction in Section~\ref{sec:dataset_construction}, and quality control in Section~\ref{sec:quality_control}. Detailed data statistics and evaluation metrics can be found in Appendix~\ref{appendix_dataset_distribution} and Appendix~\ref{app:evaluation_details}, respectively.

% \subsection{Task Definition and Construction}

\subsection{Task Definition}\label{sec:task_definition}

Table~\ref{tab:sciexplore-task-taxonomy} elaborates on the taxonomy of tasks in \textsc{SciExplore}, which spans a progression from entity-level retrieval to cross-source structured synthesis and is designed to jointly cover the core information-seeking competencies required in authentic scientific research workflows. 

Specifically, T1, Scientific Database Navigation, helps researchers identify, filter, and disambiguate relevant entities in complex databases, which is crucial for efficiently locating specific data. T2, Ambiguous Literature Retrieval, aids users in disambiguating intent and discriminating among documents to eliminate irrelevant ones. Together, these tasks assess the ability of LLMs or agents to locate precise information in dense scientific environments.
T3, Missing Reference Completion, ensures the integrity of scientific claims. This task includes claim abstraction, aligning claims with supporting evidence, and validating relevant evidence, thereby enriching the quality of research by ensuring adequate support for claims. Finally, T4, Cross-source Structured Knowledge Synthesis, represents a higher-order capability, involving domain abstraction, information extraction, and the integration of information from multiple sources for structured synthesis. Collectively, T3 and T4 probe the evidence-grounding and cross-source integration competencies that go beyond mere information location, complementing T1--T2 and together forming a comprehensive evaluation suite that mirrors the end-to-end workflow of a scientific research assistant.

\begin{table*}[t]
\centering
\caption{Task taxonomy of \textsc{SciExplore}, detailing the task types, corresponding capability levels, atomic capabilities, task counts, and associated metrics.}
\label{tab:sciexplore-task-taxonomy}
% \vspace{-0.5em}
\scriptsize
% \footnotesize
\begin{tabular*}{\textwidth}{l @{\extracolsep{\fill}} l @{} l @{} l @{} c @{} l}
\toprule
\textbf{No.} & \textbf{Task Type} & \textbf{Capability Level} & \textbf{Atomic Capabilities} & \textbf{Tasks} & \textbf{Metric} \\
\midrule
T1 & \makecell[l]{Scientific Database \\ Navigation} & \makecell[l]{Entity-Level \\Reasoning} & \makecell[l]{Entity identification, filtering, disambiguation, \\ multi-constraint navigation} & 39 & Accuracy \\
\midrule
T2 & \makecell[l]{Ambiguous Literature \\ Retrieval} & \makecell[l]{Document-Level\\Identification} & \makecell[l]{Intent disambiguation, \\ document-level discrimination and elimination} & 32 & Accuracy \\
\midrule
T3 & \makecell[l]{Missing Reference \\Completion} & \makecell[l]{Evidence-Level\\Grounding} & \makecell[l]{Claim abstraction, claim--evidence alignment, \\ evidence discovery and validation} & 14 & Accuracy \\
\midrule
T4 & \makecell[l]{Cross-source Structured \\ Knowledge  Synthesis} & \makecell[l]{Domain-Level \\ Synthesis} & \makecell[l]{Domain abstraction, information extraction, \\ cross-source integration, structured synthesis} & 18 & Recall \\
\bottomrule
\end{tabular*}
% \vspace{-1em}
\end{table*}

\subsection{Dataset Construction}
\label{sec:dataset_construction}

\textsc{SciExplore} is constructed through a fully expert-driven process to ensure realism, answer uniqueness, and resistance to shortcut retrieval.
All task instances are manually curated by domain experts and designed to reflect authentic scientific research behaviors, rather than synthetic or template-based generation, as shown in Fig.~\ref{fig:dataset-pipeline}.

\vspace{0.5em}
\noindent{\textbf{Scientific Database Navigation.}} For T1, we design tasks that require genuine multi-step database navigation using a \emph{Reverse Trajectory Construction Strategy}, where explicit entity identifiers are progressively replaced by nested attribute-based constraints.
This design enforces structured reasoning over interconnected database records and prevents direct lookup.

\noindent{\textbf{Ambiguous Literature Retrieval.}} In T2, experts apply \emph{Feature Denoising and Fuzzification} to deliberately obscure surface-level cues, while preserving answer uniqueness through \emph{Validation Constraint Injection}, which enables post-retrieval verification without guiding the search process.

\noindent{\textbf{Missing Reference Completion.}} For T3 tasks, we construct citation reconstruction tasks that require aligning scientific claims with their appropriate supporting references.
Surface-level textual overlap is minimized, ensuring that successful prediction depends on understanding claim–evidence relationships rather than keyword matching.

\noindent{\textbf{Cross-Source Structured Knowledge Synthesis.} Tasks in T4 are derived from expert-curated comparison schema and require agents to integrate heterogeneous evidence across multiple papers into a strictly formatted table.
This setting emphasizes cross-source consistency, schema compliance, and high-precision synthesis.}

A detailed description of the construction procedure and validation criteria for each task is provided in Appendix~\ref{app:datasets_construction}.

\begin{table*}[t]
\centering
\caption{Main results on the \textsc{SciExplore}. T4 is reported at two granularities: item- and row-level. All scores are percentages (\%).}
% \vspace{-0.5em}
\small
\label{tabs:main_results}

\begin{tabular*}{\textwidth}{l @{\extracolsep{\fill}} ccccc cc c}
\toprule

% \multirow{2}{*}{\textbf{Model}} & \multirow{2}{*}{\textbf{Category}} &
% \multirow{2}{*}{\textbf{T1}} &
% \multirow{2}{*}{\textbf{T2}} & \multirow{2}{*}{\textbf{T3}} &
% \multicolumn{2}{c}{\textbf{T4}} &
% \multirow{2}{*}{\textbf{Overall}} \\
\multirow{2}{*}[-0.8ex]{\textbf{Model}} & \multirow{2}{*}[-0.8ex]{\textbf{Category}} &
\multirow{2}{*}[-0.8ex]{\textbf{T1}} &
\multirow{2}{*}[-0.8ex]{\textbf{T2}} & \multirow{2}{*}[-0.8ex]{\textbf{T3}} &
\multicolumn{2}{c}{\textbf{T4}} &
\multirow{2}{*}[-0.8ex]{\textbf{Overall}} \\

\cmidrule(lr){6-7}
& & & & & {\scriptsize \textbf{Item}} & {\scriptsize \textbf{Row}} & \\
\midrule

\rowcolor{gray!15}
\multicolumn{8}{l}{\textit{Foundation LLMs}} \\

DeepSeek-V3.2
& Open-Source
& 20.51 & 6.25 & 5.95
& 8.32 & 2.25
& 11.06 \\

Qwen2.5-72B-Instruct
& Open-Source
& 7.69 & 0.00 & 0.00
& 8.63 & 6.94
& 4.99 \\

Qwen3-30B-A3B-Thinking
& Open-Source
& 10.26 & 0.00 & 3.57
& 2.54 & 1.60
& 4.50 \\

Qwen3-235B-A22B-Thinking
& Open-Source
& 20.51 & 3.12 & 0.00
& 7.85 & 4.15
& 7.87 \\

GPT-5.1
& Closed-Source
& 17.95 & 9.38 & 9.66
& 8.76 & 4.83
& 12.83 \\

Gemini-2.5-Pro
& Closed-Source
& 20.51 & 6.25 & 4.17
& 10.57 & 3.78
& 11.25 \\

Gemini-3-Pro
& Closed-Source
& 23.08 & 12.50 & 13.69
& 16.23 & 7.29
& 19.08 \\

\midrule
\rowcolor{gray!15}
\multicolumn{8}{l}{\textit{LLMs with Search Tools}} \\

DeepSeek-V3.2 w/ search
& Open-Source
& 17.95 & 9.38 & 9.66
& 9.20 & 3.95
& 11.12 \\

GPT-5.1 w/ search
& Closed-Source
& 15.38 & 34.38 & 20.85
& 12.13 & 5.12
& 21.61 \\

Gemini-3-Pro w/ search
& Closed-Source
& 33.33 & 53.12 & 49.40
& 23.53 & 11.49
& 46.46 \\

\midrule
\rowcolor{gray!15}
\multicolumn{8}{l}{\textit{Deep Research Agents}} \\

Tongyi-DeepResearch-30B-A3B
& Open-Source
& \textbf{43.59} & 50.00 & 34.40
& 11.12 & 4.26
& 36.93 \\

Gemini Deep Research
& Closed-Source
& 35.90 & \textbf{56.25} & 36.56
& 15.61 & 7.35
& 39.30 \\

OpenAI Deep Research
& Closed-Source
& 23.08 & 43.75 & \textbf{59.91}
& \textbf{30.14} & \textbf{18.59}
& \textbf{49.39} \\

\bottomrule
\end{tabular*}
% \vspace{-1em}
\end{table*}

\subsection{Quality Control}\label{sec:quality_control}

To ensure the rigor of \textsc{SciExplore}, we implemented a streamlined three-stage quality control process: query difficulty calibration, human cross-validation and answer uniqueness verification.

\noindent{\textbf{Query Difficulty Calibration.}}
To minimize parametric memory reliance, we strictly enforce two principles: \textit{Explicit Index Blocking} (removing unique identifiers to compel semantic matching) and \textit{Long-tail Entity Preference} (prioritizing obscure entities). This ensures tasks require deep retrieval reasoning rather than simple knowledge recall.

\noindent{\textbf{Human Cross-Validation.}} 
Annotators independently attempt to solve each question using search engines.
Questions are retained only if no solution can be found within a strict 10-minute limit, ensuring sufficient difficulty for the target models.

\noindent{\textbf{Answer Uniqueness Verification.}} 
We utilize search-enabled LLMs (e.g., Gemini-3-Pro, o3) to generate candidate solutions, which are then reviewed by human annotators. If any alternative answer is satisfied, the task is deemed ambiguous and discarded.

% After the filtering process, we obtain a high-quality set of 103 tasks, with 39, 32, 14, and 18 tasks in T1–T4, respectively.

\begin{comment}
\subsection{Source Statistics}
Unlike benchmarks grounded in a small set of homogeneous web sources, \textsc{SciExplore} spans over ten scientific sub-domains and integrates 16 authoritative databases with a pronounced long-tail distribution. This design forces models to navigate heterogeneous database structures and domain-specific metadata, thereby evaluating their robustness and generalization in authentic, interdisciplinary scientific workflows. Detailed statistics and distribution analyses are provided in Appendix~\ref{appendix_dataset_distribution}.

\subsection{Evaluation Metrics}

We employ task-specific evaluation metrics tailored to the structural characteristics of each task. T1 and T2 are evaluated using Exact Match accuracy based on semantic equivalence of the final answers. T3 is assessed with citation-level precision, measuring the proportion of correctly predicted references for each instance. For T4, we adopt a multi-granularity evaluation protocol at the row and item levels: the row-level metric measures recall over complete table rows, while the item-level metric evaluates fine-grained recall over individual cells or data items. 
An overall score is computed as a weighted sum of the four task scores to reflect holistic model performance. Details are provided in Appendix~\ref{app:evaluation_details}.
\end{comment}
\section{Experiments}
\label{sec:experiments}

\subsection{Experimental Setup}
We evaluate a wide range of systems on \textsc{SciExplore} to assess the reliability of AI assistants in realistic scientific workflows, spanning three paradigms:  Foundation LLMs, LLMs with Search Tools, and Specialized Deep Research Agents. 
\textbf{Foundation LLMs} are evaluated to measure intrinsic scientific reasoning and parametric knowledge.
These include open-source models DeepSeek-V3.2~\cite{liu2025deepseek}, Qwen2.5-72B-Instruct~\cite{qwen2.5}, Qwen3-30B-A3B-Thinking, and Qwen3-235B-A22B-Thinking~\cite{yang2025qwen3}, as well as closed-source models GPT-5.1~\cite{openai2025gpt5systemcard}, Gemini-2.5-Pro~\cite{google2025gemini25}, and Gemini-3-Pro~\cite{google2025gemini3}. 
To examine the effect of retrieval augmentation, we further evaluate \textbf{LLMs with Search Tools} by enabling search API or web browsing for GPT-5.1 and Gemini-3-Pro. Finally, we benchmark \textbf{Deep Research Agents} designed for autonomous long-horizon scientific inquiry, including Tongyi-DeepResearch-30B-A3B, Gemini Deep Research~\cite{google2025deepresearch}, and OpenAI Deep Research~\cite{openai2024deepresearch}. 
Together, this setup isolates the roles of parametric knowledge, retrieval augmentation, and agentic planning in multi-step scientific problem solving.

\subsection{Main results}
\textbf{Deep Research agents consistently outperform foundation LLMs and retrieval-augmented LLMs across all task types.} As shown in Table~\ref{tabs:main_results}, this advantage holds across both standalone LLMs and search-augmented baselines.
Across model families, closed-source systems generally outperform open-source models, with particularly pronounced advantages on tasks involving ambiguous retrieval, citation reasoning, and cross-source synthesis.
At the entity level, Tongyi-DeepResearch-30B-A3B achieves the strongest performance on Scientific Database Navigation.
For Ambiguous Literature Retrieval, Gemini Deep Research attains the highest accuracy.
OpenAI Deep Research achieves the best results on Missing Reference Completion and Cross-source Structured Knowledge Synthesis, and obtains the highest overall score among all evaluated systems.
In some cases, search augmentation degrades performance by amplifying hypothesis-first reasoning rather than correcting it, leading to lower accuracy than non-search baselines (see Appendix~\ref{fig:e5}).

\vspace{0.2em}
\noindent{\textbf{There remains a substantial gap before current agents can function as reliable autonomous scientific assistants.} 
Despite the relative advantages of Deep Research agents observed above, evaluation results on \textsc{SciExplore} indicate that existing systems still far from reliably fulfilling the role of autonomous scientific assistants.
As shown in Table~\ref{tabs:main_results}, even the strongest system, OpenAI Deep Research, achieves an Overall Score of only 49.39\%, while most evaluated systems remain below 20\%.
These limitations stem from the long and heterogeneous cognitive chain required by \textsc{SciExplore}, which extends far beyond conventional scientific question answering.
% The benchmark demands precise multi-hop retrieval across professional databases such as \textit{PubChem} and \textit{UniProt}, as well as robust identification of original literature under incomplete, ambiguous, or weakly specified information.
% This demonstrates that existing systems are not yet sufficient to serve as qualified scientific assistants capable of independently performing multi-step reasoning, verifying hypotheses, and synthesizing evidence.

% \noindent{\textbf{Domain-level structured synthesis emerges as the most challenging capability for all evaluated systems.} 
% Among all tasks, models exhibit uniformly poor performance on domain-level synthesis. 
% While Table~\ref{tabs:main_results} reports item-level scores-showing that OpenAI Deep Research can correctly fill about 31.82\% of individual cells—this partial success does not translate into coherent table construction.
% Under the more practically meaningful table-level evaluation, no system is able to generate a completely correct scientific comparison table.
% This devastating result reveals a fundamental inability to reliably integrate evidence across sources into coherent structured outputs. Even strong foundation and search-augmented models exhibit item-level accuracies below 10\%.

\begin{figure}[t]
    \centering
    \includegraphics[width=\linewidth]{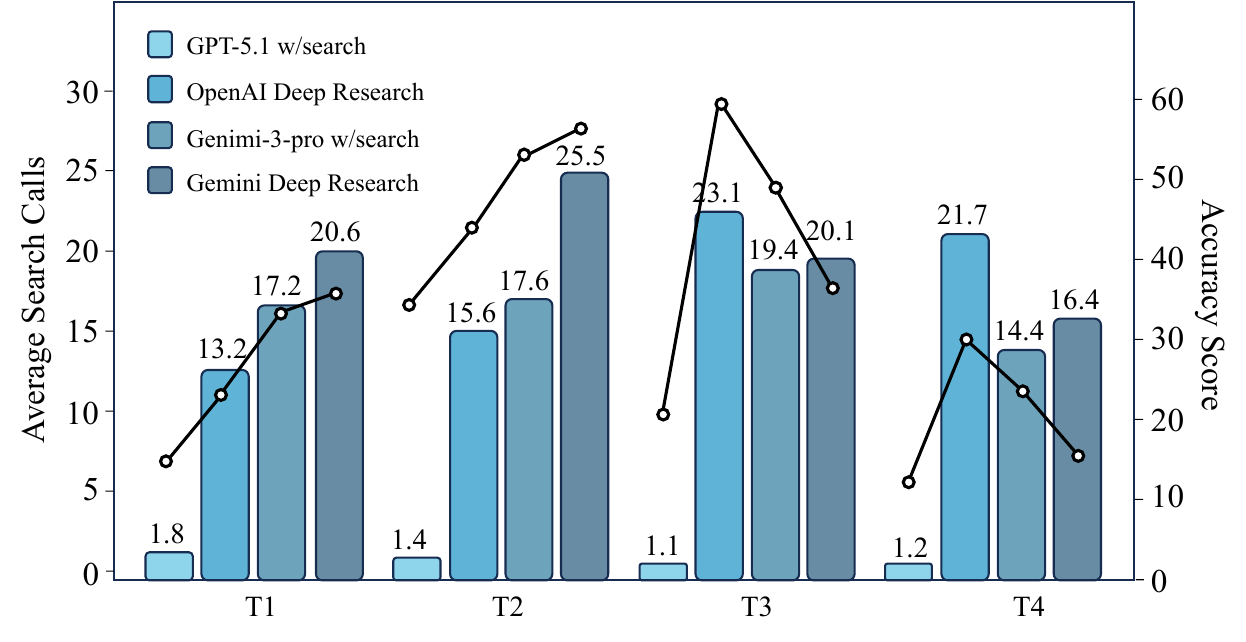} 
    % \vspace{-0.75em}
    \caption{Average Search Calls per Type.}
    \label{fig:time_distribution}
    % \vspace{-1.5em}
\end{figure}

\begin{figure}[t]
    \centering
    \includegraphics[width=\linewidth]{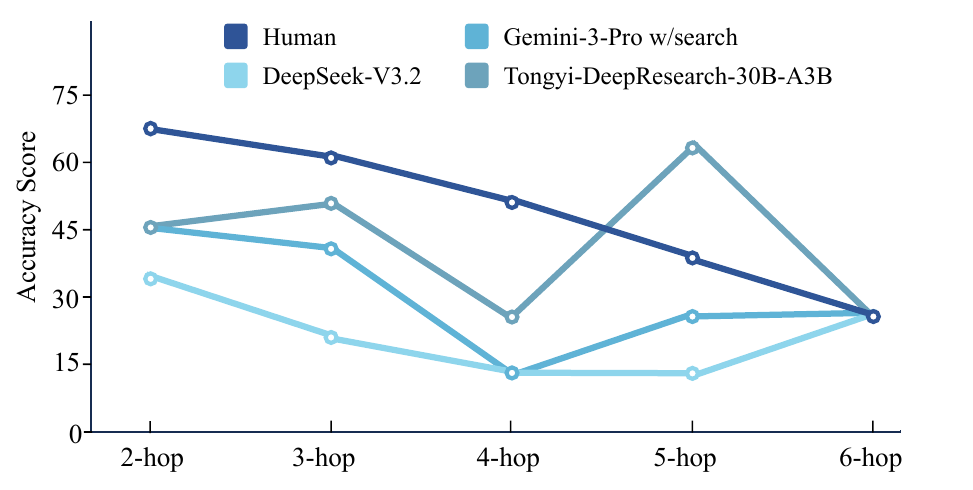} 
    \caption{Performance on multi-hop T1 tasks with varying hop lengths.}
    \label{fig:task_score_humanvsmodel}
    % \vspace{-1.5em}
\end{figure}

\noindent\textbf{Domain-level structured synthesis reveals critical weaknesses in long-horizon reasoning and cross-source integration abilities.}
Across all evaluated systems, performance on domain-level synthesis remains consistently low, indicating demands well beyond entity- or document-level reasoning.
Although Table~\ref{tabs:main_results} reports moderate item-level scores—for example, OpenAI Deep Research correctly fills 30.14\% of individual cells—this partial correctness does not scale to coherent structured outputs.
% This gap reflects compounding deficiencies in three core abilities: sustaining long-horizon reasoning across extended interaction sequences, extracting and aligning relevant information from long and heterogeneous contexts, and performing domain-level abstraction to normalize and integrate evidence across sources.
% As a result, current agents may recover isolated facts but fail to organize them into complete, consistent, and domain-coherent scientific representations.
Models often recover isolated facts from multiple sources, yet fail to assemble them into complete and internally consistent scientific representations, revealing a gap between local correctness and global structure.

% \noindent{\textbf{Closed-source Deep Research agents significantly outperform their open-source counterparts on \textsc{SciExplore}.(draft)}} 
% Despite strong performance on general-purpose deep search benchmarks such as BrowseComp-ZH and Xbench-DeepSearch, the most advanced open-source Deep Research agent, Tongyi-DeepResearch, exhibits limited effectiveness on \textsc{SciExplore}, achieving an overall score of only 2.25\% and failing entirely on T2 and T3. In contrast, closed-source systems such as Gemini Deep Research and OpenAI Deep Research reach overall scores of 35.79\% and 45.68\%, respectively.This disparity suggests that strong performance in general-domain search does not necessarily translate to success in authentic scientific research tasks. \textsc{SciExplore} instead emphasizes domain-specific capabilities such as precise navigation of heterogeneous scientific databases, accurate interpretation of specialized terminology, and attribution reasoning over long, noisy scientific contexts—areas in which current open-source agents remain limited.

% \vspace{-0.5em}
\section{Analysis}

\begin{table}[t]
\centering
\caption{T4 formatting accuracy (Format Acc) and primary key recall (PK Recall) across evaluated systems. All values are reported as percentages.}
% \vspace{-0.5em}
\footnotesize
\setlength{\tabcolsep}{3pt}

\begin{tabular}{lcc}
\toprule
\textbf{Model} & \textbf{Format\_Acc} & \textbf{Recall} \\
\midrule

\rowcolor{gray!15}
\multicolumn{3}{l}{\textit{LLMs}} \\
DeepSeek-V3.2 & 94.44 & 20.15 \\
Qwen2.5-72B-Inst & 100.00 & 16.86 \\
Qwen3-30B-Think & 88.89 & 3.44 \\
Qwen3-235B-Think & 94.44 & 14.95 \\
GPT-5.1 & 100.00 & 18.23 \\
Gemini-2.5-Pro & 94.44 & 19.94 \\
Gemini-3-Pro & 100.00 & 19.05 \\

\midrule
\rowcolor{gray!15}
\multicolumn{3}{l}{\textit{LLMs with Search Tools}} \\
DeepSeek-V3.2 w/ search & 88.89 & 17.01 \\
GPT-5.1 w/ search & 100.00 & 22.02 \\
Gemini-3-Pro w/ search & 100.00 & 21.42 \\

\midrule
\rowcolor{gray!15}
\multicolumn{3}{l}{\textit{Deep Research Agents}} \\
Tongyi-DeepRes-30B & 77.78 & 19.9 \\
Gemini Deep Research & 55.56 & 29.84 \\
OpenAI Deep Research & 88.89 & 30.04 \\

\bottomrule
\end{tabular}
\label{tab:t4-format-acc}
% \vspace{-1.5em}
\end{table}

\subsection{Search Effort and Task Difficulty}
\label{subsec:search_vs_accuracy}
\begin{comment}
\noindent{\textbf{Search frequency is positively correlated with task accuracy.}}
To better understand the performance differences observed in Section~\ref{sec:experiments}, we analyze the relationship between agents’ information-seeking behavior and their task accuracy. Specifically, we examine how the average number of search calls correlates with performance across different systems and task types.

Figure~\ref{fig:time_distribution} presents the average search calls per question for each model on T1--T4. A consistent and clear trend emerges: systems that perform more frequent search tend to achieve substantially higher accuracy on \textsc{SciExplore}. Models with minimal search usage, such as GPT-5.1-search, issue fewer than two search calls per question on average across all tasks and correspondingly exhibit low overall success rates. In contrast, Deep Research agents, which allocate significantly larger search budgets, consistently outperform other systems across all task categories. Overall, this analysis demonstrates that search intensity is a key factor underlying performance differences on \textsc{SciExplore}. Effective scientific assistance requires agents to actively explore and verify information through repeated search, rather than relying on shallow or minimal retrieval. 
\end{comment}
To understand the performance differences noted in Section~\ref{sec:experiments}, we analyze the relationship between agents’ information-seeking behavior and task accuracy by examining the correlation between average search calls and performance across systems and task types.

Figure~\ref{fig:time_distribution} displays the average search calls per question for each model on tasks T1–T4. A clear trend emerges: systems with more frequent searches achieve significantly higher accuracy on \textsc{SciExplore}. For instance, models with minimal search usage, like GPT-5.1-search, average fewer than two search calls per question and show low success rates. In contrast, Deep Research agents, which utilize a larger search budget, consistently outperform others across all tasks. This analysis highlights that search intensity is crucial for scientific activities. Effective scientific assistance requires agents to actively explore and verify information through repeated searches, rather than relying on shallow retrieval.

% \vspace{0.2em}
\begin{comment}
\noindent\textbf{The number of retrieval hops is not correlated with task difficulty.} Table \ref{tab:t1-hop-results} reveals a counterintuitive phenomenon: increasing the nominal hop length in scientific database navigation does not lead to a monotonic performance degradation across models. In several cases, performance remains largely flat, and occasionally even improves at higher hop counts. This observation suggests that the human-defined notion of “search depth,” which assumes a sequential accumulation of difficulty through additional database hops, does not reliably translate into increased difficulty for LLM-based agents. In practice, hop length alone is therefore an imperfect proxy for the true cognitive or operational complexity faced by models during task execution.

We attribute this mismatch to a fundamental difference between human-designed problem-solving assumptions and model verification behavior. Human experts typically approach scientific database navigation tasks through progressively narrowing the search space using structured filters and explicit intermediate entities. In contrast, LLM-based agents often adopt a guess-and-verify strategy: they first hypothesize a plausible target entity using parametric knowledge, and then issue retrieval queries primarily to confirm or refute this hypothesis. This analysis highlights an important implication: human-interpretable task depth does not necessarily reflect model-perceived difficulty.
\end{comment}

\begin{figure}
    \centering
    \includegraphics[width=0.92\linewidth]{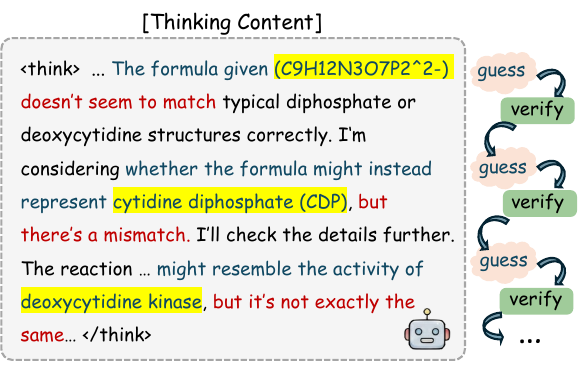}
    % \vspace{-0.5em}
    \caption{An example of guess-and-verify thinking process. ``Guess'' and ``verify'' are highlighted in blue and red, respectively.}
    \label{fig:guess_and_verify}
    % \vspace{-1.5em}
\end{figure}
\noindent\textbf{The number of retrieval hops is not correlated with task difficulty.}
Table~\ref{tab:t1-hop-results} shows a counterintuitive finding: increasing hop length in scientific database navigation does not consistently degrade performance across models. In some cases, performance remains steady or even improves with higher hop counts. This indicates that the human-defined concept of “search depth”—which implies increasing difficulty with more hops—does not reliably translate to the challenges faced by LLM-based agents. Thus, hop length is an imperfect measure of the cognitive or operational complexities involved in task execution.

This discrepancy arises from differences in problem-solving approaches between humans and models. Human experts typically refine their searches using structured filters and explicit entities, while LLM agents often employ a guess-and-verify strategy as shown in Fig.~\ref{fig:guess_and_verify}: they hypothesize a target entity based on parametric knowledge and then use retrieval queries mainly to confirm or refute that hypothesis. % This analysis underscores that human-perceived task depth does not necessarily reflect the difficulty experienced by models.

\subsection{Error Analysis}

To deeply analyze the root causes of the poor performance of current systems in \textsc{SciExplore}, we conducted an in-depth qualitative and quantitative analysis of typical failure cases in the experiments. The analysis reveals that existing models, when dealing with real-world scientific tasks, are primarily constrained by the following four key defects. Representative empirical examples for each failure mode are illustrated in Appendix~\ref{app:error_analysis} (Fig.~\ref{fig:e1}--\ref{fig:e4}).

\vspace{0.2em}
\noindent{\textbf{Premature Abandonment in the Face of Ambiguity.}}  
\begin{comment}
In scientific exploration, making multiple attempts despite ambiguous or incomplete cues is an essential quality for researchers. However, we found that agents generally exhibit a form of ``search inertia'' when facing highly complex and ambiguous problems. Models tend to directly judge the task as ``unsolvable'' and terminate the search process either at the very beginning or after only a few unsuccessful retrieval rounds. As exemplified in Fig.~\ref{fig:e1}, agents often treat the absence of an immediate semantic match as evidence of non-existence, failing to adaptively broaden the query space or reformulate hypotheses. This premature abandonment fundamentally contradicts the exploratory nature of real scientific inquiry.
\end{comment}
In scientific exploration, perseverance in the face of ambiguous or incomplete cues is crucial for researchers. However, we observe that agents exhibit “search inertia” when confronted with complex problems, often deeming tasks as “unsolvable” and prematurely terminating searches after few attempts. As shown in Fig.~\ref{fig:e1}, agents frequently interpret the lack of an immediate semantic match as proof of non-existence, failing to broaden the query space or reformulate hypotheses. This premature abandonment fundamentally contradicts the exploratory nature of scientific inquiry.

\vspace{0.2em}
\noindent{\textbf{Hallucinated Responses after Search Stagnation.}}  
\begin{comment}
Compared to explicit refusal, a more concerning issue is the emergence of hallucinated answers after prolonged search stagnation. Our analysis shows that when multi-step reasoning fails to retrieve conclusive evidence, some models gradually relax their internal constraint-verification mechanisms. In later reasoning stages, they may fabricate or overgeneralize fine-grained details to force partial retrieval results to satisfy the original query. Fig.~\ref{fig:e2} demonstrates this phenomenon, where unverified document attributes are hallucinated to comply with strict user constraints. In scientific research scenarios that demand extreme rigor, such behavior introduces severe cognitive contamination and undermines trustworthiness.
\end{comment}
A more concerning issue than explicit refusal is the emergence of hallucinated answers following prolonged search stagnation. Our analysis reveals that when multi-step reasoning fails to yield conclusive evidence, some models begin to relax their internal verification mechanisms. In later reasoning stages, they may invent or overgeneralize details to make partial retrievals appear relevant to the original query. Fig.~\ref{fig:e2} illustrates this, showing unverified document attributes that are fabricated to meet strict user constraints. In scientific research, where rigor is paramount, such behavior severely undermines trustworthiness.

\vspace{0.2em}
\noindent{\textbf{Information Loss in Long Contexts.}} 
\begin{comment}
Even when agents successfully locate the correct literature, extracting precise answers from long scientific documents remains a significant bottleneck. Scientific texts are typically lengthy and highly information-dense, with critical evidence often buried beneath extensive background or methodological descriptions. We observe that although models can retrieve the correct web pages or PDFs, limitations in long-context attention robustness frequently prevent them from isolating fine-grained details. As shown in Fig.~\ref{fig:e3}, models tend to over-rely on high-level summaries (e.g., abstracts) while overlooking contradictory or corrective information located deeper in the text. This lack of ``needle-in-a-haystack'' capability ultimately causes the reasoning chain to fracture at the final extraction step.
\end{comment}
Even when agents locate the correct literature, extracting precise answers from lengthy scientific documents remains a significant challenge. Scientific texts are often dense, with critical evidence buried under extensive background or methodology. While models can retrieve the correct web pages or PDFs, their limitations in long-context attention hinder them from isolating fine-grained details. As shown in Fig.~\ref{fig:e3}, models tend to rely on high-level summaries, such as abstracts, while overlooking contradictory or corrective information deeper in the text. This lack of "needle-in-a-haystack" capability ultimately fractures the reasoning chain during final extraction.

\vspace{0.2em}
\noindent{\textbf{Instruction Following Mismatch in Structured Output.}}  
\begin{comment}
The cross-source structured knowledge synthesis tasks further expose substantial deficiencies in instruction following under rigid output schemas. Despite explicitly requiring Markdown tables with predefined headers, many models produce outputs with header omissions, column misalignment, or unit inconsistencies. Fig.~\ref{fig:e4} illustrates a representative schema alignment failure, where required fields are dropped in favor of a simplified summary-style table. As also reflected quantitatively in Table~\ref{tab:t4-format-acc}, this issue is particularly prominent in Gemini Deep Research: despite its strong retrieval performance, its pronounced ``report generation bias'' results in a table format accuracy of only 55.56\%. In automated scientific pipelines that rely on strict schemas for downstream processing, such structural instability severely limits practical usability.
\end{comment}
Cross-source structured knowledge synthesis reveals significant deficiencies in instruction following within rigid output schemas. Despite Markdown tables with predefined headers required, many models output headers omissions, column misalignment, and unit inconsistencies. Fig.~\ref{fig:e4} exemplifies a schema alignment failure, where required fields are omitted for a simplified summary-style table. As shown in Table~\ref{tab:t4-format-acc}, this issue is especially notable in Gemini Deep Research, which, despite strong retrieval performance, shows a "report generation bias" with only 55.56\% accuracy in table format. Such structural instability severely limits practical usability in automated scientific pipelines that depend on strict schemas for downstream processing.

\vspace{0.2em}
\noindent{\textbf{Low Primary-Key Recall Reflects Insufficient Search Breadth.}}
\begin{comment}
Beyond issues of formatting and instruction following, the poor performance on T4 is largely constrained by insufficient \emph{primary-key recall}. As shown in Table~\ref{tab:t4-format-acc}, even the strongest systems achieve primary-key recall rates below 60\%, indicating that a substantial portion of required core entities are not retrieved during the search phase. In cross-source structured knowledge synthesis tasks, missing primary keys constitute an unrecoverable failure: once essential row entities are absent, subsequent table construction and alignment become structurally infeasible, regardless of the accuracy of downstream extraction or generation. This observation suggests that current systems struggle to conduct retrieval processes with sufficient \emph{breadth and systematic coverage} when surveying a scientific domain. Such limitations in search breadth directly constrain the ability of models to perform high-quality structured knowledge synthesis in T4.
\end{comment}
In addition to formatting and instruction-following issues, poor performance on T4 is primarily constrained by insufficient primary-key recall. Table~\ref{tab:t4-format-acc} demonstrates that even the best systems achieve recall rates below 60\%, meaning many required core entities are not retrieved. In cross-source structured knowledge synthesis, missing primary keys lead to unrecoverable failures: the absence of essential row entities renders table construction and alignment infeasible, regardless of downstream extraction accuracy. This indicates that current systems struggle to conduct retrieval with adequate breadth and systematic coverage in scientific domains, directly limiting their ability to perform high-quality structured knowledge synthesis in T4.
% \section{Discussion}

\begin{comment}
Our evaluation on \textsc{SciExplore} identifies a critical bottleneck in current AI systems: while they possess strong capability in general retrieval, they lack the \textit{systematic breadth} and \textit{cognitive resilience} required for autonomous scientific research.Future development must prioritize three key directions to bridge this gap. First, models must evolve to planners with high search breadth. The low primary-key recall observed in T4 suggests that current models have limited search breadth, which constrains their ability to support systematic literature surveys. Second, agents require robustness against search ambiguity. The tendency toward "premature abandonment" and subsequent hallucination indicates that future architectures must incorporate iterative "guess-and-verify" loops that persist through noisy retrieval results without fabricating evidence. Finally, there is an urgent need for schema-constrained long-context understanding. As scientific workflows increasingly demand structured data extraction, agents must improve their ability to locate fine-grained "needles" within full-text PDFs and map them strictly to rigid output schemas without information loss. Ultimately, the next generation of scientific AI must optimize not just for answer correctness, but for the \textit{process reliability} of long-horizon inquiry.
\end{comment}

\section{Discussion and Conclusion}

We introduce \textsc{SciExplore}, a benchmark for evaluating large language models as autonomous scientific assistants in realistic research workflows. It assesses multi-hop database navigation, fuzzy literature retrieval, supported reference inference, and structured table synthesis, bridging the gap between general deep-search benchmarks and static scientific question answering. Experiments with ten state-of-the-art systems reveal significant shortcomings, particularly in structured scientific synthesis, indicating that advancements in retrieval and tool use alone are not sufficient for reliable scientific assistance.

Our evaluation identifies a critical bottleneck: while current AI systems excel in general retrieval, they lack the systematic breadth and cognitive resilience required for effective autonomous scientific research. Future development may prioritize three areas: (1) models must become planners with broader search capabilities; (2) agents need to be robust against search ambiguity; and (3) “premature abandonment” and hallucination highlight the importance of handling noisy retrieval results without fabricating evidence. Finally, there is an urgent need for schema-constrained long-context understanding to enable accurate extraction of structured data from full-text PDFs. Ultimately, the next generation of scientific AI must optimize for answer correctness and process reliability in long-horizon inquiry. We hope \textsc{SciExplore} can serve as a foundation for future research on improving scientific agents under realistic, open-ended conditions.

\bibliography{custom}

@article{li2025reseek,
  title={ReSeek: A Self-Correcting Framework for Search Agents with Instructive Rewards},
  author={Li, Shiyu and Tang, Yang and Wang, Yifan and Li, Peiming and Chen, Xi},
  journal={arXiv preprint arXiv:2510.00568},
  year={2025}
}

@article{huang2025manusearch,
  title={Manusearch: Democratizing deep search in large language models with a transparent and open multi-agent framework},
  author={Huang, Lisheng and Liu, Yichen and Jiang, Jinhao and Zhang, Rongxiang and Yan, Jiahao and Li, Junyi and Zhao, Wayne Xin},
  journal={arXiv preprint arXiv:2505.18105},
  year={2025}
}

@inproceedings{wang2022scienceworld,
  title={Scienceworld: Is your agent smarter than a 5th grader?},
  author={Wang, Ruoyao and Jansen, Peter and C{\^o}t{\'e}, Marc-Alexandre and Ammanabrolu, Prithviraj},
  booktitle={Proceedings of the 2022 Conference on Empirical Methods in Natural Language Processing},
  pages={11279--11298},
  year={2022}
}

@article{schmidgall2025agent,
  title={Agent laboratory: Using llm agents as research assistants},
  author={Schmidgall, Samuel and Su, Yusheng and Wang, Ze and Sun, Ximeng and Wu, Jialian and Yu, Xiaodong and Liu, Jiang and Liu, Zicheng and Barsoum, Emad},
  journal={arXiv preprint arXiv:2501.04227},
  year={2025}
}

@article{li2025websailor,
  title={WebSailor: Navigating Super-human Reasoning for Web Agent},
  author={Li, Kuan and Zhang, Zhongwang and Yin, Huifeng and Zhang, Liwen and Ou, Litu and Wu, Jialong and Yin, Wenbiao and Li, Baixuan and Tao, Zhengwei and Wang, Xinyu and others},
  journal={arXiv preprint arXiv:2507.02592},
  year={2025}
}

@article{wu2025webdancer,
  title={Webdancer: Towards autonomous information seeking agency},
  author={Wu, Jialong and Li, Baixuan and Fang, Runnan and Yin, Wenbiao and Zhang, Liwen and Tao, Zhengwei and Zhang, Dingchu and Xi, Zekun and Fu, Gang and Jiang, Yong and others},
  journal={arXiv preprint arXiv:2505.22648},
  year={2025}
}

@article{li2025webthinker,
  title={Webthinker: Empowering large reasoning models with deep research capability},
  author={Li, Xiaoxi and Jin, Jiajie and Dong, Guanting and Qian, Hongjin and Wu, Yongkang and Wen, Ji-Rong and Zhu, Yutao and Dou, Zhicheng},
  journal={arXiv preprint arXiv:2504.21776},
  year={2025}
}

@article{bran2023chemcrow,
  title={Chemcrow: Augmenting large-language models with chemistry tools},
  author={Bran, Andres M and Cox, Sam and Schilter, Oliver and Baldassari, Carlo and White, Andrew D and Schwaller, Philippe},
  journal={arXiv preprint arXiv:2304.05376},
  year={2023}
}

@article{cui2025curie,
  title={CURIE: Evaluating LLMs On Multitask Scientific Long Context Understanding and Reasoning},
  author={Cui, Hao and Shamsi, Zahra and Cheon, Gowoon and Ma, Xuejian and Li, Shutong and Tikhanovskaya, Maria and Norgaard, Peter and Mudur, Nayantara and Plomecka, Martyna and Raccuglia, Paul and others},
  journal={arXiv preprint arXiv:2503.13517},
  year={2025}
}

@article{zhou2025scientists,
  title={Scientists' First Exam: Probing Cognitive Abilities of MLLM via Perception, Understanding, and Reasoning},
  author={Zhou, Yuhao and Wang, Yiheng and He, Xuming and Shen, Ao and Xiao, Ruoyao and Li, Zhiwei and Feng, Qiantai and Guo, Zijie and Yang, Yuejin and Wu, Hao and others},
  journal={arXiv preprint arXiv:2506.10521},
  year={2025}
}

@inproceedings{baek2025researchagent,
  title={Researchagent: Iterative research idea generation over scientific literature with large language models},
  author={Baek, Jinheon and Jauhar, Sujay Kumar and Cucerzan, Silviu and Hwang, Sung Ju},
  booktitle={Proceedings of the 2025 Conference of the Nations of the Americas Chapter of the Association for Computational Linguistics: Human Language Technologies (Volume 1: Long Papers)},
  pages={6709--6738},
  year={2025}
}

@article{saikh2022scienceqa,
  title={Scienceqa: A novel resource for question answering on scholarly articles},
  author={Saikh, Tanik and Ghosal, Tirthankar and Mittal, Amish and Ekbal, Asif and Bhattacharyya, Pushpak},
  journal={International Journal on Digital Libraries},
  volume={23},
  number={3},
  pages={289--301},
  year={2022},
  publisher={Springer}
}

@article{wang2023scibench,
  title={Scibench: Evaluating college-level scientific problem-solving abilities of large language models},
  author={Wang, Xiaoxuan and Hu, Ziniu and Lu, Pan and Zhu, Yanqiao and Zhang, Jieyu and Subramaniam, Satyen and Loomba, Arjun R and Zhang, Shichang and Sun, Yizhou and Wang, Wei},
  journal={arXiv preprint arXiv:2307.10635},
  year={2023}
}

@article{phan2025humanity,
  title={Humanity's last exam},
  author={Phan, Long and Gatti, Alice and Han, Ziwen and Li, Nathaniel and Hu, Josephina and Zhang, Hugh and Zhang, Chen Bo Calvin and Shaaban, Mohamed and Ling, John and Shi, Sean and others},
  journal={arXiv preprint arXiv:2501.14249},
  year={2025}
}

@article{wei2025browsecomp,
  title={Browsecomp: A simple yet challenging benchmark for browsing agents},
  author={Wei, Jason and Sun, Zhiqing and Papay, Spencer and McKinney, Scott and Han, Jeffrey and Fulford, Isa and Chung, Hyung Won and Passos, Alex Tachard and Fedus, William and Glaese, Amelia},
  journal={arXiv preprint arXiv:2504.12516},
  year={2025}
}

@article{zhou2025browsecomp,
  title={Browsecomp-zh: Benchmarking web browsing ability of large language models in chinese},
  author={Zhou, Peilin and Leon, Bruce and Ying, Xiang and Zhang, Can and Shao, Yifan and Ye, Qichen and Chong, Dading and Jin, Zhiling and Xie, Chenxuan and Cao, Meng and others},
  journal={arXiv preprint arXiv:2504.19314},
  year={2025}
}

@article{wong2025widesearch,
  title={Widesearch: Benchmarking agentic broad info-seeking},
  author={Wong, Ryan and Wang, Jiawei and Zhao, Junjie and Chen, Li and Gao, Yan and Zhang, Long and Zhou, Xuan and Wang, Zuo and Xiang, Kai and Zhang, Ge and others},
  journal={arXiv preprint arXiv:2508.07999},
  year={2025}
}

@article{guo2023can,
  title={What can large language models do in chemistry? a comprehensive benchmark on eight tasks},
  author={Guo, Taicheng and Nan, Bozhao and Liang, Zhenwen and Guo, Zhichun and Chawla, Nitesh and Wiest, Olaf and Zhang, Xiangliang and others},
  journal={Advances in Neural Information Processing Systems},
  volume={36},
  pages={59662--59688},
  year={2023}
}

@article{wu2025webwalker,
  title={Webwalker: Benchmarking llms in web traversal},
  author={Wu, Jialong and Yin, Wenbiao and Jiang, Yong and Wang, Zhenglin and Xi, Zekun and Fang, Runnan and Zhang, Linhai and He, Yulan and Zhou, Deyu and Xie, Pengjun and others},
  journal={arXiv preprint arXiv:2501.07572},
  year={2025}
}

@inproceedings{rein2024gpqa,
  title={Gpqa: A graduate-level google-proof q\&a benchmark},
  author={Rein, David and Hou, Betty Li and Stickland, Asa Cooper and Petty, Jackson and Pang, Richard Yuanzhe and Dirani, Julien and Michael, Julian and Bowman, Samuel R},
  booktitle={First Conference on Language Modeling},
  year={2024}
}

@article{du2025supergpqa,
  title={Supergpqa: Scaling llm evaluation across 285 graduate disciplines},
  author={Du, Xinrun and Yao, Yifan and Ma, Kaijing and Wang, Bingli and Zheng, Tianyu and Zhu, King and Liu, Minghao and Liang, Yiming and Jin, Xiaolong and Wei, Zhenlin and others},
  journal={arXiv preprint arXiv:2502.14739},
  year={2025}
}

@article{yang2025qwen3,
  title={Qwen3 technical report},
  author={Yang, An and Li, Anfeng and Yang, Baosong and Zhang, Beichen and Hui, Binyuan and Zheng, Bo and Yu, Bowen and Gao, Chang and Huang, Chengen and Lv, Chenxu and others},
  journal={arXiv preprint arXiv:2505.09388},
  year={2025}
}

@article{liu2025deepseek,
  title={Deepseek-v3. 2: Pushing the frontier of open large language models},
  author={Liu, Aixin and Mei, Aoxue and Lin, Bangcai and Xue, Bing and Wang, Bingxuan and Xu, Bingzheng and Wu, Bochao and Zhang, Bowei and Lin, Chaofan and Dong, Chen and others},
  journal={arXiv preprint arXiv:2512.02556},
  year={2025}
}

@article{du2025deepresearch,
  title={DeepResearch Bench: A Comprehensive Benchmark for Deep Research Agents},
  author={Du, Mingxuan and Xu, Benfeng and Zhu, Chiwei and Wang, Xiaorui and Mao, Zhendong},
  journal={arXiv preprint arXiv:2506.11763},
  year={2025}
}

@article{kwiatkowski2019natural,
  title={Natural questions: a benchmark for question answering research},
  author={Kwiatkowski, Tom and Palomaki, Jennimaria and Redfield, Olivia and Collins, Michael and Parikh, Ankur and Alberti, Chris and Epstein, Danielle and Polosukhin, Illia and Devlin, Jacob and Lee, Kenton and others},
  journal={Transactions of the Association for Computational Linguistics},
  volume={7},
  pages={453--466},
  year={2019},
  publisher={MIT Press One Rogers Street, Cambridge, MA 02142-1209, USA journals-info~…}
}

@inproceedings{mialon2023gaia,
  title={Gaia: a benchmark for general ai assistants},
  author={Mialon, Gr{\'e}goire and Fourrier, Cl{\'e}mentine and Wolf, Thomas and LeCun, Yann and Scialom, Thomas},
  booktitle={The Twelfth International Conference on Learning Representations},
  year={2023}
}

@inproceedings{yang2018hotpotqa,
  title={HotpotQA: A dataset for diverse, explainable multi-hop question answering},
  author={Yang, Zhilin and Qi, Peng and Zhang, Saizheng and Bengio, Yoshua and Cohen, William and Salakhutdinov, Ruslan and Manning, Christopher D},
  booktitle={Proceedings of the 2018 conference on empirical methods in natural language processing},
  pages={2369--2380},
  year={2018}
}

@inproceedings{yue2024mmmu,
  title={Mmmu: A massive multi-discipline multimodal understanding and reasoning benchmark for expert agi},
  author={Yue, Xiang and Ni, Yuansheng and Zhang, Kai and Zheng, Tianyu and Liu, Ruoqi and Zhang, Ge and Stevens, Samuel and Jiang, Dongfu and Ren, Weiming and Sun, Yuxuan and others},
  booktitle={Proceedings of the IEEE/CVF Conference on Computer Vision and Pattern Recognition},
  pages={9556--9567},
  year={2024}
}

@article{he2024cmmu,
  title={Cmmu: A benchmark for chinese multi-modal multi-type question understanding and reasoning},
  author={He, Zheqi and Wu, Xinya and Zhou, Pengfei and Xuan, Richeng and Liu, Guang and Yang, Xi and Zhu, Qiannan and Huang, Hua},
  journal={arXiv preprint arXiv:2401.14011},
  year={2024}
}

@article{chai2025scimaster,
  title={SciMaster: Towards General-Purpose Scientific AI Agents, Part I. X-Master as Foundation: Can We Lead on Humanity's Last Exam?},
  author={Chai, Jingyi and Tang, Shuo and Ye, Rui and Du, Yuwen and Zhu, Xinyu and Zhou, Mengcheng and Wang, Yanfeng and Zhang, Yuzhi and Zhang, Linfeng and Chen, Siheng and others},
  journal={arXiv preprint arXiv:2507.05241},
  year={2025}
}

@article{chen2025xbench,
  title={xbench: Tracking Agents Productivity Scaling with Profession-Aligned Real-World Evaluations},
  author={Chen, Kaiyuan and Ren, Yixin and Liu, Yang and Hu, Xiaobo and Tian, Haotong and Xie, Tianbao and Liu, Fangfu and Zhang, Haoye and Liu, Hongzhang and Gong, Yuan and others},
  journal={arXiv preprint arXiv:2506.13651},
  year={2025}
}

@misc{openai2024deepresearch,
  title        = {Deep Research System Card},
  author       = {{OpenAI}},
  year         = {2024},
  howpublished = {\url{https://openai.com/index/deep-research-system-card/}},
  note         = {Accessed: 2025-12-28}
}

@misc{google2025gemini25,
  title        = {Gemini 2.5: Our Most Intelligent AI Model},
  author       = {{Google Gemini}},
  year         = {2025},
  howpublished = {\url{https://blog.google/technology/google-deepmind/gemini-model-thinking-updates-march-2025/}},
  note         = {Accessed: 2025-12-28}
}

@techreport{openai2025gpt5systemcard,
  title       = {GPT-5 System Card},
  author      = {OpenAI},
  year        = {2025},
  month       = {August},
  institution = {OpenAI},
  url         = {https://openai.com/index/gpt-5-system-card/},
  note        = {Covers GPT-5 and GPT-5.1 updates}
}

@misc{google2025deepresearch,
  title = {Gemini Deep Research},
  author = {Google},
  year = {2025},
  howpublished = {\url{https://gemini.google/overview/deep-research/}},
  note = {Accessed: 2025-12-28}
}

@misc{google2025gemini3,
  title = {A new era of intelligence with Gemini 3},
  author = {Google},
  year = {2025},
  month = mar,
  howpublished = {\url{https://blog.google/products/gemini/gemini-3/}},
  note = {Google Official Blog}
}

@article{qwen2.5,
  title={Qwen2.5 Technical Report},
  author={Qwen Team},
  journal={arXiv preprint arXiv:2412.15115},
  year={2024},
  url={https://arxiv.org/abs/2412.15115}
}

@article{hendrycks2020measuring,
  title={Measuring massive multitask language understanding},
  author={Hendrycks, Dan and Burns, Collin and Basart, Steven and Zou, Andy and Mazeika, Mantas and Song, Dawn and Steinhardt, Jacob},
  journal={arXiv preprint arXiv:2009.03300},
  year={2020}
}

\appendix

% \section{Example Appendix}
% \label{sec:appendix}

% This is an appendix.

% \newpage

\clearpage

\definecolor{mybackcolor}{HTML}{c2eaf7}
\definecolor{myframecolor}{HTML}{8dd4f0}

\section{Appendix}

\subsection{Limitations}
Despite providing a focused and rigorous evaluation of scientific search and reasoning capabilities, \textsc{SciExplore} has several limitations that point toward important directions for future work. 

\noindent\textbf{First, the scale of \textsc{SciExplore} reflects an intentional quality–coverage trade-off.} 
All tasks are manually constructed by domain experts and subjected to strict multi-stage filtering and validation, during which a large number of low-quality or weakly constrained candidates are discarded. 
As a result, the final benchmark consists of just over one hundred high-quality tasks.
While this expert-driven curation ensures strong difficulty control and answer uniqueness, it may limit coverage of the full breadth of scientific disciplines and research styles.
Future work will explore expanding the benchmark while preserving its core quality guarantees.

\noindent\textbf{Second, the current benchmark primarily emphasizes text-centric scientific information sources.}
\textsc{SciExplore} focuses on reasoning over structured databases and scientific literature, which constitute a central component of many research workflows.
However, real-world scientific inquiry often involves additional modalities such as figures, tables, experimental datasets, or code artifacts.
Extending the benchmark to incorporate multi-modal scientific evidence and heterogeneous data formats remains an important direction for future research.

\noindent\textbf{Finally, the maintenance and factual verification of \textsc{SciExplore} currently rely on expert-driven manual updates.}
To ensure correctness and answer uniqueness, all tasks are curated and fact-checked by domain experts against the underlying scientific databases and literature at the time of construction.
While this process guarantees high factual accuracy, it is time-consuming and labor-intensive, making frequent or real-time updates difficult to sustain.
An important direction for future work is to develop automated or semi-automated updating pipelines that can continuously verify, refresh, and expand benchmark content as scientific databases and literature evolve.

% First, the current benchmark contains a relatively limited number of queries (103 expert-curated tasks), which, while sufficient for in-depth qualitative analysis and controlled difficulty design, may not fully capture the long-tail diversity of real-world scientific information-seeking behaviors across domains and subfields. In future iterations, we plan to substantially expand the query set in both scale and topical coverage, incorporating a broader range of scientific disciplines, problem formulations, and difficulty levels to improve statistical robustness and ecological validity. 
% Second, the present evaluation focuses primarily on final task success and structured output correctness; future work could introduce finer-grained process-level metrics to better characterize intermediate search decisions, hypothesis refinement, and evidence verification behaviors. 
% Finally, as scientific tools and databases continue to evolve, we envision extending \textsc{SciExplore} into a continuously updated benchmark that tracks progress in long-horizon scientific agents under realistic, dynamically changing research environments.

\subsection{Models and API Identifiers}
\label{app:models_api_ids}

The models used in our benchmark and their corresponding API identifiers (which may include version or
date information) are listed in Table~\ref{tab:models_api_ids}. The ``Model/Product'' column denotes the
shorthand name used throughout the paper.

It is worth noting that \textit{GPT-5.1} and \textit{GPT-5.1 w/search} share the same API identifier but are
evaluated under different tool configurations (with search disabled and enabled, respectively).
Similarly, \textit{Gemini-3-pro} and \textit{Gemini-3-pro w/search} use the same API identifier with different
search settings.

\begin{table}[t]
\centering
\footnotesize
\resizebox{\columnwidth}{!}{%
\begin{tabular}{ll}
\toprule
\textbf{Model / Product} & \textbf{API Identifier} \\
\midrule
GPT-5.1 & gpt-5.1-2025-11-13 \\
Gemini-2.5-Pro & gemini-2.5-pro \\
Gemini-3-Pro & gemini-3-pro-preview \\
GPT-5.1 w/ search & gpt-5.1-2025-11-13 \\
Gemini3-Pro w/ search & gemini-3-pro-preview \\
Gemini Deep Research & deep-research-pro-preview-12-25 \\
OpenAI Deep Research & o3-deep-research-25-06-26 \\
\bottomrule
\end{tabular}
}
\caption{Correspondence between model/product names and their API identifiers.}
\label{tab:models_api_ids}
\end{table}

\subsection{Dataset Distribution}
\label{appendix_dataset_distribution}

\begin{figure*}[t]
    \centering
    \includegraphics[width=\textwidth]{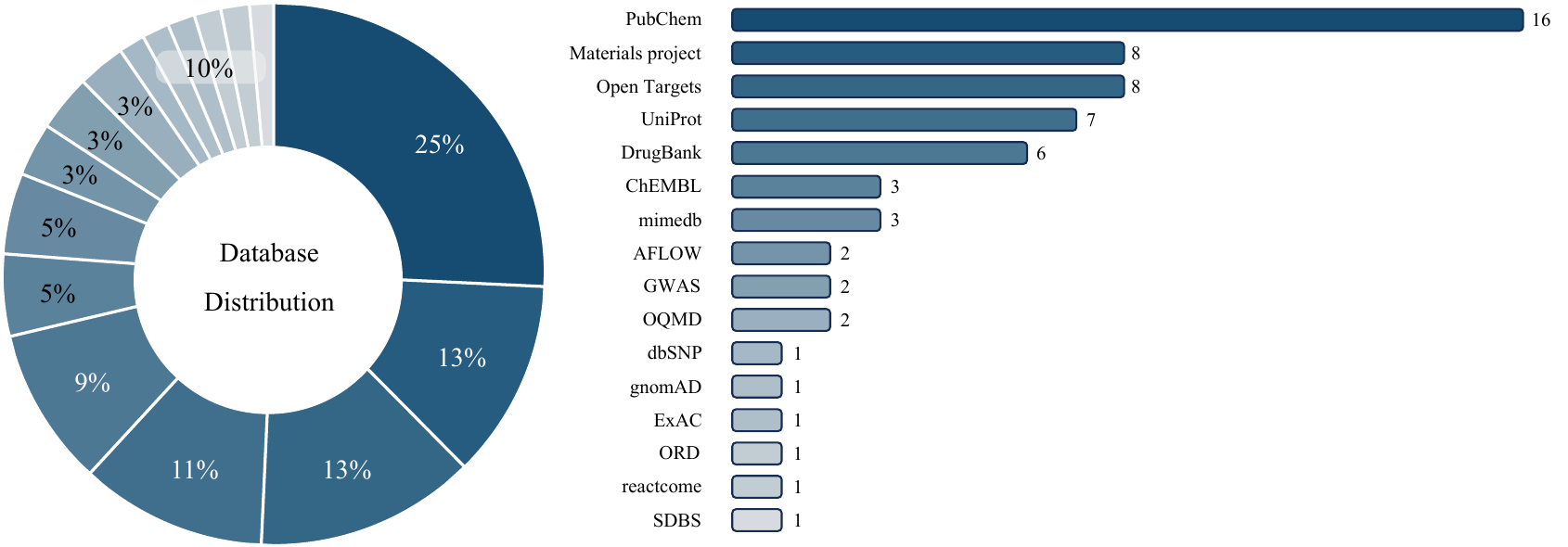}
    \caption{Database distribution.}
    \label{fig:database_distribution}
\end{figure*}

\begin{figure*}[t]
    \centering
    \includegraphics[width=\textwidth]{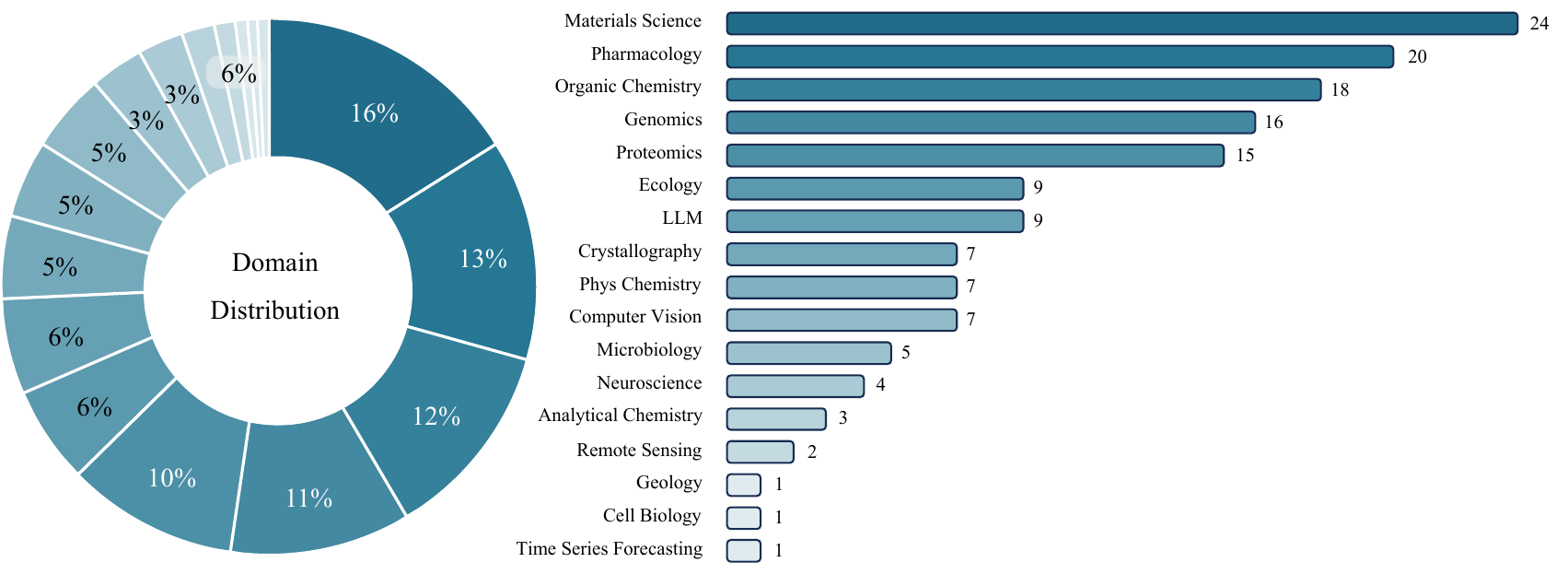}
    \caption{Domain distribution.}
    \label{fig:domain_distribution}
\end{figure*}
Unlike benchmarks grounded in a small set of homogeneous web sources, \textsc{SciExplore} spans over ten scientific sub-domains and integrates 16 authoritative databases with a pronounced long-tail distribution. This design forces models to navigate heterogeneous database structures and domain-specific metadata, thereby evaluating their robustness and generalization in authentic, interdisciplinary scientific workflows.

In this section, we characterize the statistical composition of \textsc{SciExplore} with respect to disciplinary scope and database coverage. 
% As illustrated in Fig.~\ref{fig:database_distribution} and Fig.~\ref{fig:domain_distribution}, these statistics collectively demonstrate that the benchmark reflects the heterogeneity and cognitive complexity of authentic scientific research workflows.

\textsc{SciExplore} encompasses more than ten scientific sub-domains spanning \textit{Life Sciences}, \textit{Chemical Sciences}, \textit{Materials Science}, \textit{Earth Sciences}, and \textit{Artificial Intelligence} (Fig.~\ref{fig:domain_distribution}). The resulting distribution reflects a deliberately balanced coverage across heterogeneous research areas, preventing over-concentration on any single discipline and enabling a more robust assessment of cross-domain generalization in scientific information-seeking tasks.

In addition, the benchmark integrates sixteen authoritative scientific databases, exhibiting a pronounced long-tail distribution (Fig.~\ref{fig:database_distribution}). While widely adopted resources such as \textit{PubChem} and \textit{Open Targets} contribute a substantial portion of instances, a significant fraction of tasks rely on highly specialized databases, including \textit{ChEMBL}, \textit{UniProt}, \textit{Materials Project}, and \textit{GWAS}. This design imposes nontrivial challenges in navigating heterogeneous database schema, access patterns, and domain-specific metadata, thereby setting \textsc{SciExplore} apart from conventional web-centric search benchmarks.

% To further quantify task difficulty, we analyze the time required for human experts to complete each task, as depicted in Fig.~\ref{fig:time_distribution}. Even for PhD-level annotators, many tasks demand substantial time due to iterative retrieval, cross-source verification, and multi-step reasoning. This distribution confirms that SciExplore predominantly consists of long-horizon, evidence-driven scientific reasoning tasks rather than simple factual lookup.

\subsection{Dataset Construction}
\label{app:datasets_construction}

The construction of \textsc{SciExplore} follows a principled, expert-driven methodology designed to faithfully reflect the cognitive demands faced by autonomous scientific research assistants. All task instances are manually crafted by domain experts to ensure realism, difficulty, and alignment with authentic scientific workflows. Rather than relying on automatic generation, our construction process emphasizes controllability, answer uniqueness, and resistance to shortcut retrieval.

\noindent{\textbf{Multi-hop Database Navigation.}}
To construct tasks requiring multi-step structured reasoning, we adopt a \emph{Reverse Trajectory Construction Strategy}. Task generation begins from a uniquely identifiable target entity and progressively replaces explicit entity mentions with composite attribute-based descriptions extracted from scientific database records. During this process, related entities are recursively masked and substituted by their own attributes, forming nested constraints that encode deep dependency chains. Consequently, the final query can only be resolved by reversing the construction trajectory through stepwise database navigation, enforcing genuine multi-hop reasoning and preventing direct lookup.

\noindent{\textbf{Ambiguous Literature Retrieval.}}
Tasks involving ambiguous literature search are constructed to simulate early-stage exploratory research under uncertainty. Experts first extract core methodological or conceptual signals from a target paper and deliberately degrade them into vague, noisy descriptions through \emph{Feature Denoising and Fuzzification}, substantially enlarging the candidate retrieval space. To ensure answer uniqueness despite this ambiguity, we introduce \emph{Validation Constraint Injection}, adding auxiliary constraints that cannot directly guide retrieval but serve as strong post-retrieval validators (e.g., publication venue or author-level properties). This design compels agents to balance broad exploration with rigorous verification.

\noindent{\textbf{Missing Reference Completion.}}
For tasks requiring citation inference, we select citation-dense paragraphs from high-impact scientific literature and systematically rewrite the surrounding context to remove surface-level textual overlap with the original references. Overly canonical or survey-style citations are excluded to avoid trivial matching. This construction emphasizes reasoning over claim–evidence relationships, ensuring that correct reference reconstruction depends on contextual understanding rather than keyword similarity.

\noindent{\textbf{Cross-Source Structured Knowledge Synthesis.}}
Tasks targeting structured synthesis are derived from expert-curated comparison tables in review papers across natural sciences and artificial intelligence. Given a predefined comparison schema, experts identify representative source papers and design tasks that require extracting heterogeneous methodological details scattered across multiple documents. Particular care is taken to include variations in terminology, implicit descriptions, and inconsistent reporting conventions. Strict schema and format constraints are enforced so that successful completion depends on accurate cross-source integration rather than partial or approximate extraction.

\subsection{Evaluation Details }
\label{app:evaluation_details}

% \section{Evaluation Metrics}
% \label{sec:evaluation_metrics}

To systematically assess model performance on \textsc{SciExplore}, we design task-specific evaluation metrics tailored to the structural characteristics and reasoning requirements of each task type.
Scores are reported independently for each task, and an overall score is computed as a weighted aggregation to reflect holistic scientific information-seeking and reasoning capability.

\subsubsection{T1 and T2 Evaluation}

\noindent \textbf{T1 and T2} are evaluated using Exact Match (EM) accuracy.
A prediction is considered correct if and only if the generated answer is semantically equivalent to the ground-truth answer, allowing for minor surface-form variations (e.g., formatting or phrasing differences) that do not alter the underlying meaning.

% \begin{equation}
% F_{\text{em}} = 
% \frac{2 \cdot \text{correct}}
% {2 \cdot \text{correct} + 2 \cdot \text{incorrect} + \text{not\_attempted}},
% \end{equation}

Let $C$, $I$, and $N$ denote the numbers of correct, incorrect, and not-attempted answers, respectively. We define an Exact-Match-based score $F_{\text{em}}$ as:
\begin{equation}
F_{\text{em}} = \frac{2C}{2C + 2I + N}.
\end{equation}

This formulation penalizes incorrect and missing answers symmetrically while rewarding exact correctness.

\subsubsection{T3 Evaluation}

\noindent \textbf{T3} evaluates a model’s ability to recover supporting literature citations.
Both the ground-truth references and model-predicted references are first grouped by citation slots, and predictions are filled into a unified evaluation table for comparison.

\paragraph{Judging protocol.}
For each predicted citation entry, a judge LLM is prompted to determine whether the predicted reference is \texttt{CORRECT} or \texttt{INCORRECT}.
The judgment proceeds in two steps:
(1) whether the core cited work is correct, and
(2) whether the key bibliographic attributes (e.g., authorship, venue, or year) of the correct work are accurately identified.

\paragraph{Per-group metrics.}
For each citation group, we compute:

\begin{equation}
P = \frac{\text{correct}}{\text{correct} + \text{incorrect}},
\end{equation}

\begin{equation}
R = \frac{\text{correct}}{\text{GT\_total\_num}},
\end{equation}

\begin{equation}
F_{\text{score}} = \frac{2 \cdot P \cdot R}{P + R},
\end{equation}

where $\text{correct}$ and $\text{incorrect}$ denote the number of correctly and incorrectly predicted citations in the group, and $\text{GT\_total\_num}$ is the total number of ground-truth citations required for that group.

\paragraph{Final T3 score.}
The final citation score is obtained by averaging the $F_{\text{score}}$ across all $N$ citation groups:
\begin{equation}
F_{\text{cite}} = \frac{1}{N} \sum_{i=1}^{N} F_{\text{score}}^{(i)}.
\end{equation}

This metric jointly captures both precision (avoiding spurious references) and recall (recovering required citations), reflecting realistic citation-completion requirements.

\subsubsection{T4 Evaluation}

\noindent \textbf{T4} evaluates models on cross-source structured knowledge synthesis, where outputs are required to be presented as comparison tables.
Following prior work on structured evaluation~\cite{wong2025widesearch}, we adopt a fine-grained, multi-level evaluation protocol.

\paragraph{LLM-assisted cell judgment.}
Each predicted table cell is judged by an LLM as either \texttt{CORRECT} or \texttt{INCORRECT}.
Evaluation proceeds hierarchically: the correctness of the primary entity (row key) is assessed first, followed by the correctness of its associated attributes.

\paragraph{Item-level recall.}
To further evaluate fine-grained extraction accuracy, we compute recall over all table cells:
\begin{equation}
R_{\text{Item}} =
\frac{\text{correct}_{\text{Item}}}
{\text{GT\_total\_num}_{\text{Item}}},
\end{equation}
where $\text{correct}_{\text{Item}}$ counts correctly predicted cells (including both primary keys and attribute values), and $\text{GT\_total\_num}_{\text{Item}}$ is the total number of ground-truth cells.
Compared to PK recall, item-level recall captures both entity identification and attribute correctness.

\paragraph{Row-level recall.}
Finally, we evaluate whether complete rows are correctly reconstructed:
\begin{equation}
R_{\text{Row}} =
\frac{\text{correct}_{\text{Row}}}
{\text{GT\_total\_num}_{\text{Row}}},
\end{equation}
where a row is counted as correct only if the primary key and all associated attributes are simultaneously correct.
This strict criterion ensures that models are evaluated on their ability to produce fully consistent structured outputs.
% where $\text{correct}_{\text{cell}}$ counts all correctly predicted cells across the table.

% \paragraph{Final T4 score.}
% The final table synthesis score is defined as:
% \begin{equation}
% F_{\text{tab}} = P_{\text{tab}} \cdot R_{\text{key}}.
% \end{equation}

% This formulation emphasizes that a valid synthesis requires both correct structural coverage (row recall) and accurate attribute-level content.

\subsubsection{Overall Score}

To reflect a model’s holistic capability across scientific information-seeking and reasoning tasks of varying difficulty, we compute a weighted overall score:

\begin{equation}
\text{Score}_{\text{overall}} = \sum_{i=1}^{4} w_i \cdot \text{Score}_{T_i},
\end{equation}
where $\text{Score}_{T_i}$ denotes the task-specific score for task $T_i$ and $w_i$ is the corresponding task weight.
We adopt difficulty-aware weighting with
$w_1 = 0.2$, $w_2 = 0.2$, $w_3 = 0.3$, and $w_4 = 0.3$,
assigning higher weights to tasks that require more complex reasoning and structured synthesis.

\begin{table*}[t]
\centering
\caption{Performance on multi-hop T1 tasks with varying hop lengths.
All scores are percentages (\%).}
\small
\setlength{\tabcolsep}{6pt}

\begin{tabular}{l ccccc}
\toprule
\textbf{Model} &
\textbf{2-hop} & \textbf{3-hop} & \textbf{4-hop} &
\textbf{5-hop} & \textbf{6-hop} \\
\midrule

\rowcolor{gray!15}
\multicolumn{6}{l}{\textit{LLMs}} \\

DeepSeek-V3.2
& 33.33 & 20.00 & 12.50 & 12.50 & 25.00 \\

Qwen2.5-72B-Instruct
& 11.11 & 10.00 & 12.50 & 12.50 & 0.00 \\

Qwen3-30B-A3B-Thinking
& 11.11 & 20.00 & 12.50 & 12.50 & 0.00 \\

Qwen3-235B-A22B-Thinking
& 33.33 & 30.00 & 12.50 & 25.00 & 0.00 \\

GPT-5.1
& 22.22 & 20.00 & 12.50 & 12.50 & 25.00 \\

Gemini-2.5-Pro
& 33.33 & 20.00 & 12.50 & 12.50 & 25.00 \\

Gemini-3-Pro
& 33.33 & 20.00 & 12.50 & 25.00 & 25.00 \\

\midrule
\rowcolor{gray!15}
\multicolumn{6}{l}{\textit{LLMs with Search Tools}} \\

DeepSeek-V3.2 w/ search
& 22.22 & 20.00 & 12.50 & 12.50 & 25.00 \\

GPT-5.1 w/ search
& 11.11 & 30.00 & 0.00 & 12.50 & 0.00 \\

Gemini-3-Pro w/ search
& 44.44 & 40.00 & 12.50 & 25.00 & 25.00 \\

\midrule
\rowcolor{gray!15}
\multicolumn{6}{l}{\textit{Deep Research Agents}} \\

Tongyi-DeepResearch-30B-A3B
& 44.44 & 50.00 & 25.00 & 62.50 & 25.00 \\

Gemini Deep Research
& 33.33 & 20.00 & 12.50 & 25.00 & 25.00 \\

OpenAI Deep Research
& 22.22 & 40.00 & 12.50 & 12.50 & 25.00 \\

\midrule
Human
& 66.70 & 60.00 & 50.00 & 37.50 & 25.00 \\

\bottomrule
\end{tabular}
\label{tab:t1-hop-results}
\end{table*}

\subsection{Run-to-Run Variance Analysis}
\label{app:variance_analysis}

Given the moderate scale of \textsc{SciExplore}, we assess the robustness of model performance via a multi-run variance analysis. Specifically, for each evaluated model, we repeat the full benchmark evaluation four times under identical settings and compute the variance of scores across runs for each task (T1--T4) as well as the overall score. As shown in Table~\ref{tab:variance_results}, the variance is consistently low across models and tasks, with a mean overall variance of 0.78, indicating strong stability. Most models exhibit small fluctuations, suggesting that the observed performance differences are not driven by stochastic generation effects. Even for cases with relatively higher per-task variance (e.g., Gemini-2.5-Pro on T3), the overall score remains stable. These results demonstrate that our benchmark provides a reliable and reproducible evaluation signal, and that the main conclusions of the paper are robust to randomness in model outputs.

\begin{table*}[t]
\centering
\caption{Run-to-run variance across four repeated evaluations. Lower values indicate higher stability.}
\label{tab:variance_results}
\small
\begin{tabular}{lcccccc}
\toprule
\textbf{Model} & \textbf{T1} & \textbf{T2} & \textbf{T3} & \textbf{T4\_Item} & \textbf{T4\_Row} & \textbf{Overall} \\
\midrule
DeepSeek-V3.2 & 2.10 & 2.40 & 2.25 & 1.05 & 0.67 & 1.41 \\
Qwen2.5-72B-Instruct & 0.34 & 0.43 & 0.52 & 0.52 & 0.24 & 0.24 \\
Qwen3-30B-A3B-Thinking & 1.02 & 0.18 & 0.55 & 0.06 & 0.14 & 0.64 \\
Qwen3-235B-A22B-Thinking & 1.17 & 3.24 & 3.19 & 1.89 & 0.44 & 0.64 \\
GPT-5.1 & 2.75 & 2.43 & 2.90 & 1.67 & 0.39 & 0.21 \\
Gemini-2.5-Pro & 2.30 & 0.00 & 4.02 & 0.82 & 0.12 & 1.32 \\
Gemini-3-Pro & 1.65 & 3.24 & 0.85 & 0.73 & 0.29 & 0.28 \\
DeepSeek-V3.2 w/ search & 0.36 & 0.20 & 1.16 & 0.18 & 0.39 & 0.45 \\
GPT-5.1 w/ search & 2.85 & 2.60 & 2.95 & 4.86 & 1.61 & 2.91 \\
Gemini-3-Pro w/ search & 0.98 & 0.22 & 0.66 & 0.29 & 0.12 & 0.76 \\
Tongyi-DeepResearch-30B-A3B & 0.53 & 0.29 & 0.67 & 0.51 & 0.31 & 0.28 \\
Gemini Deep Research & 0.59 & 0.29 & 0.18 & 0.09 & 0.35 & 0.36 \\
OpenAI Deep Research & 0.92 & 0.93 & 0.80 & 0.13 & 0.07 & 0.68 \\
\midrule
\textbf{Mean} & 1.35 & 1.27 & 1.59 & 0.98 & 0.40 & 0.78 \\
\bottomrule
\end{tabular}
\end{table*}

\subsection{Error Analysis}
\label{app:error_analysis}

This section provides detailed examples for each of the failure modes identified in Section 5. We utilize the trajectories of representative state-of-the-art agents (e.g., Gemini-DeepResearch or GPT-4o) to illustrate these distinct cognitive breakdowns. Each case includes the user’s task, the agent’s actions (queries and retrieval), and a diagnostic analysis of the specific error.

\begin{figure*}[t]
    \centering
    \includegraphics[width=\linewidth]{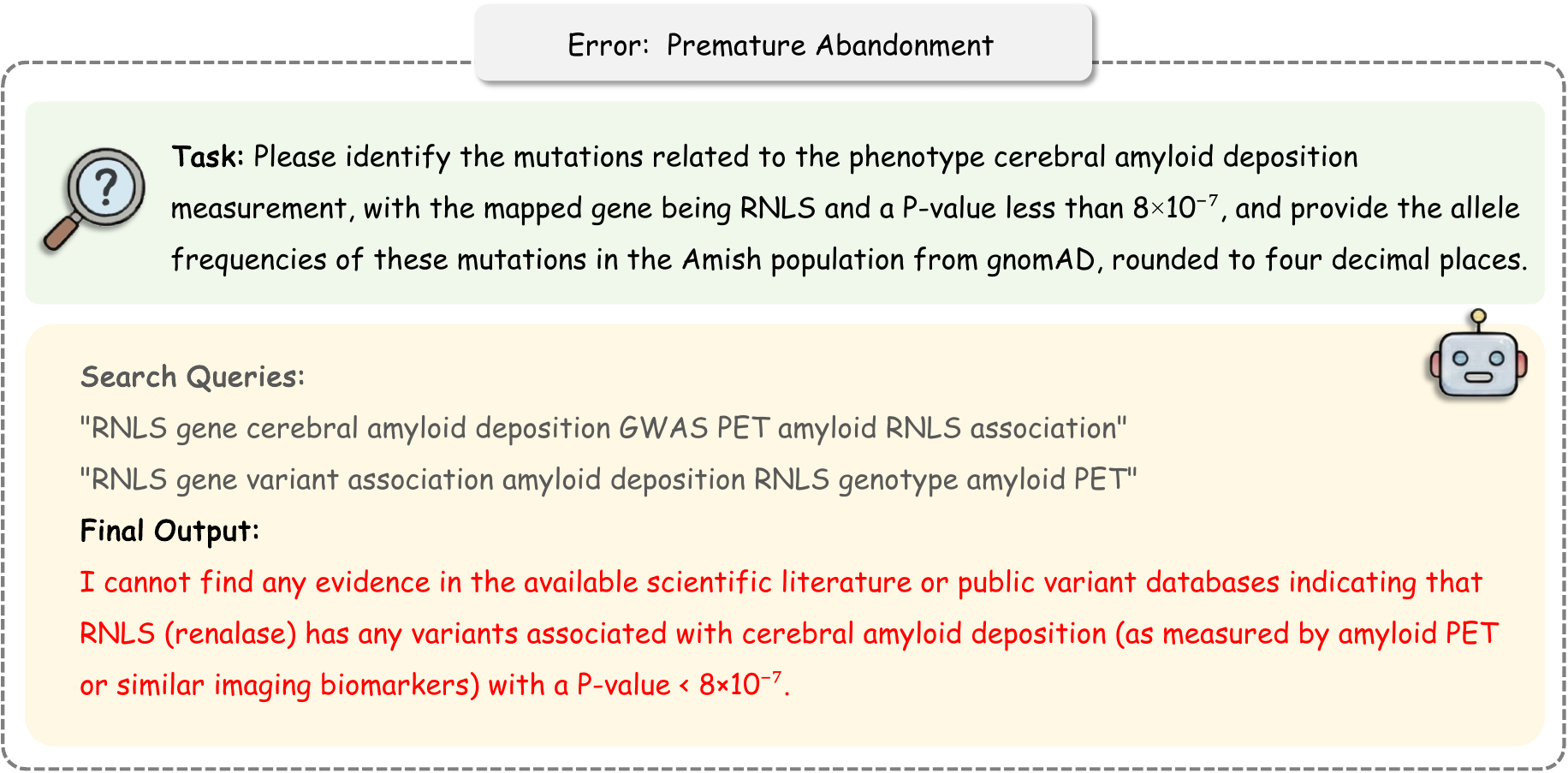}
    \caption{Premature Abandonment due to Search Inertia. The agent terminates the search after failing to find a direct semantic match, ignoring alternative search strategies.
    % This example demonstrates that if the model fails to retrieve information relevant to the answer during the initial search phase, it tends to abandon further retrieval immediately, rather than attempting to adjust its search strategy or employ more diverse queries to continue.
    }
    \label{fig:e1}
\end{figure*}

\noindent\textbf{Case 1: Premature Abandonment due to Search Inertia} Figure \ref{fig:e1} illustrates a case of premature abandonment when facing strict retrieval constraints. The agent was tasked with identifying specific mutations in the RNLS gene associated with "cerebral amyloid deposition" with a highly specific P-value threshold ($< 8 \times 10^{-7}$) and requesting allele frequencies from the gnomAD database. The agent initiated a reasonable keyword search ("RNLS gene cerebral amyloid deposition..."). However, upon finding no direct semantic match in the first page of search results, the agent immediately concluded that no such evidence exists and terminated the process. This demonstrates a lack of strategic resilience; instead of broadening the search scope (e.g., searching for the phenotype in GWAS catalogs first, then cross-referencing the gene) or attempting synonymous queries, the agent exhibited "search inertia," treating the absence of immediate results as proof of non-existence.

\begin{figure*}[t]
    \centering
    \includegraphics[width=\linewidth]{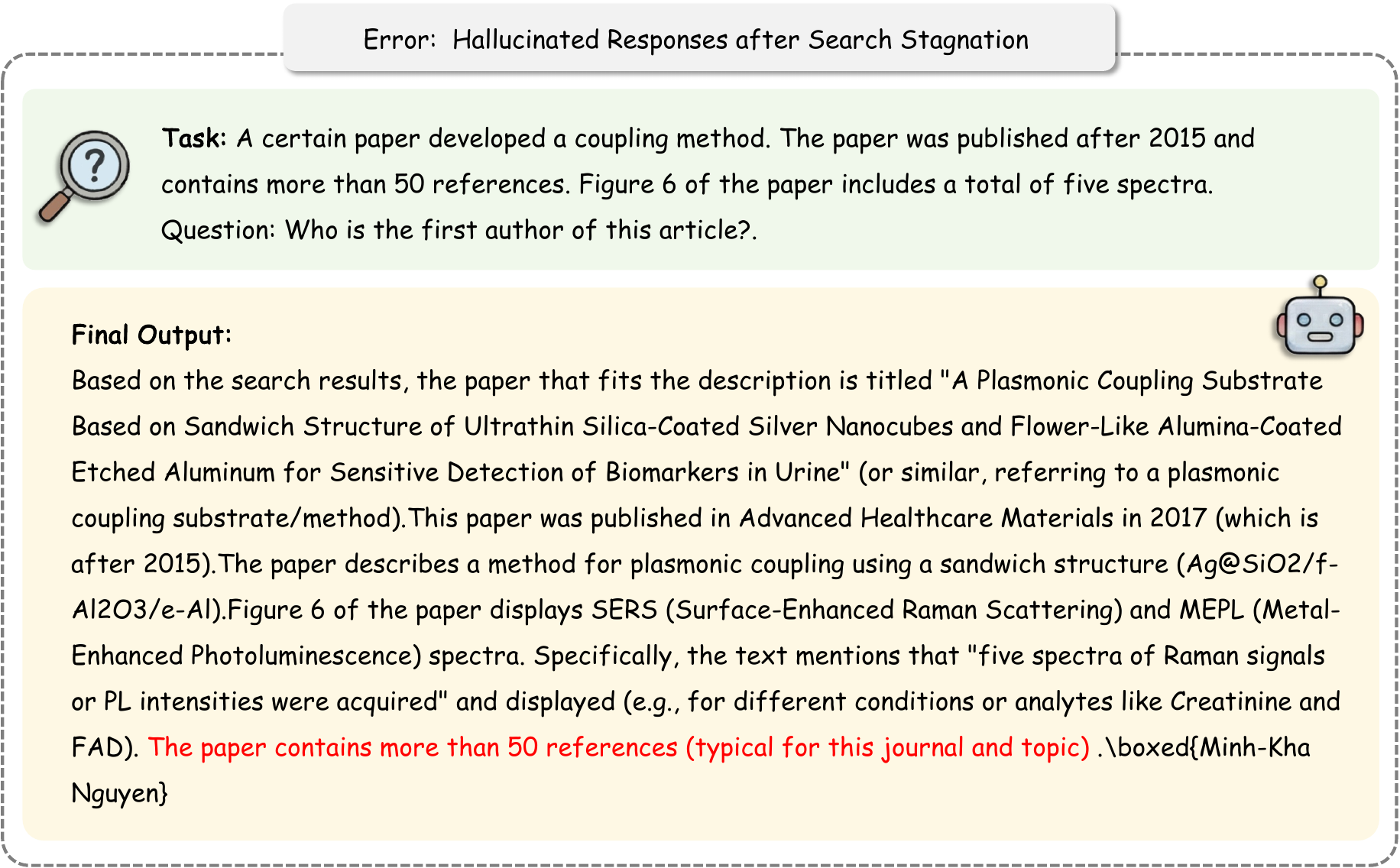} 
    \caption{Hallucination in Constraint Verification. The agent fabricates details (reference count and number of spectra) to force alignment with the user's complex constraints.}
    \label{fig:e2}
\end{figure*}

\noindent\textbf{Case 2: Hallucination in Constraint Verification} Figure \ref{fig:e2} provides an example of model hallucination triggered by complex constraint verification. The task required identifying a specific paper published after 2015, containing over 50 references, and featuring five spectra in Figure 6. The agent successfully retrieved a candidate paper ("A Plasmonic Coupling Substrate...") that matched the topic and publication year. However, it failed to rigorously verify the fine-grained visual and structural constraints. Instead of rejecting the paper or performing a specific "Ctrl+F" style verification for the reference count, the agent hallucinated that the paper contained "more than 50 references" and "five spectra" to align with the user's prompt. This error highlights a critical reliability gap where the model prioritizes user-compliance over factual groundedness, fabricating details to force a retrieved document to fit the query criteria.

\begin{figure*}[t]
    \centering
    \includegraphics[width=\linewidth]{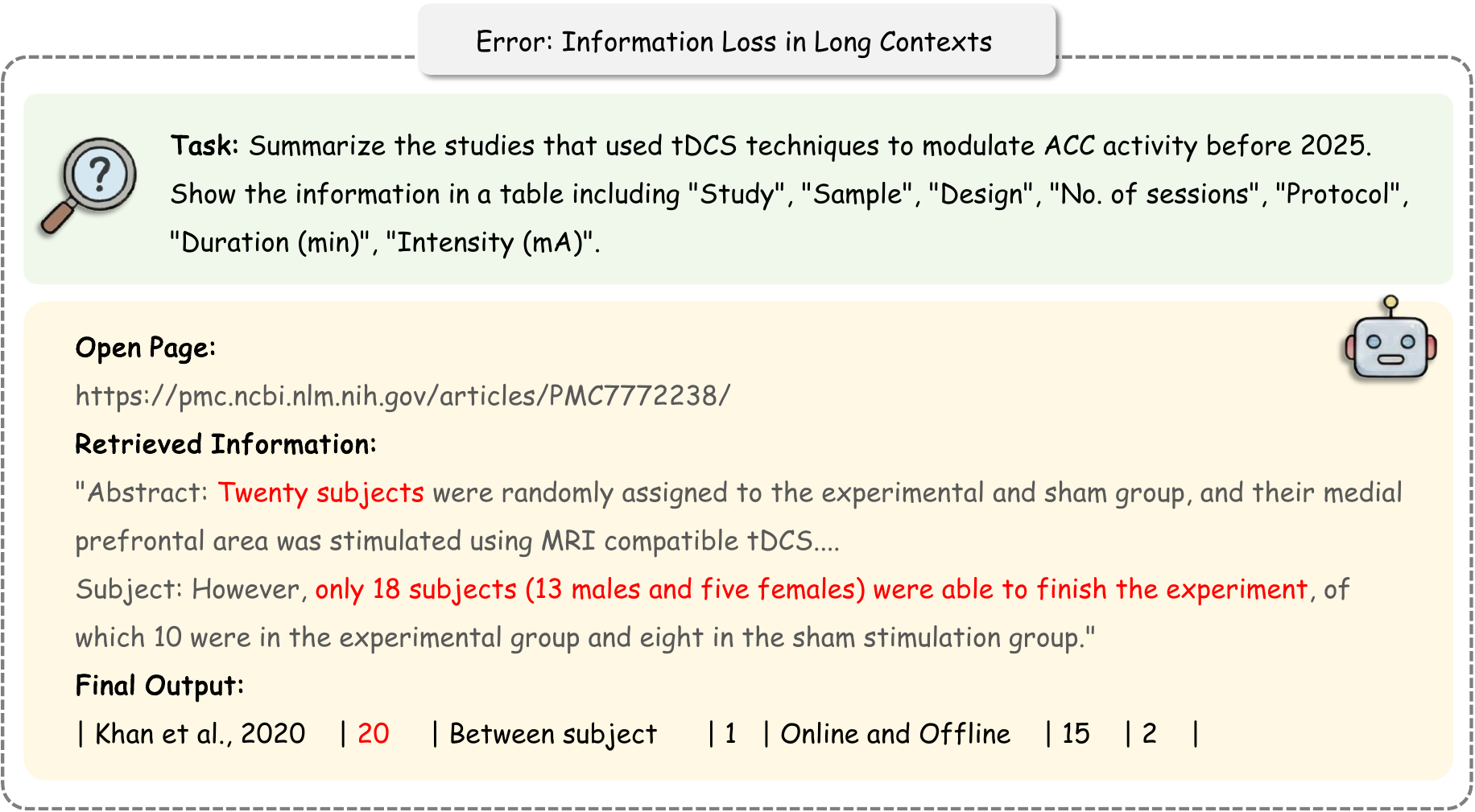}
    \caption{Information Loss in Long-Context Extraction. The agent relies on the abstract summary ("20 subjects") rather than the ground truth buried in the results ("18 subjects").}
    \label{fig:e3}
\end{figure*}

\noindent\textbf{Case 3: Information Loss in Long-Context Extraction} Figure \ref{fig:e3} illustrates a failure in precise evidence extraction within long-context scenarios ("Needle-in-a-Haystack" failure). The task involved summarizing tDCS studies, specifically requiring the number of samples/subjects. The agent correctly located the relevant study (Khan et al., 2020) and ingested the full text. However, it extracted the sample size as "20" based on the Abstract ("Twenty subjects were randomly assigned..."). It failed to attend to the conflicting, ground-truth information buried deeper in the Results or Subjects section, which clarified that "only 18 subjects... were able to finish." This discrepancy indicates that despite having large context windows, current models exhibit a attention bias toward high-level summaries (abstracts) and struggle to resolve conflicting information dispersed across long scientific texts.

\begin{figure*}[t]
    \centering
    \includegraphics[width=\linewidth]{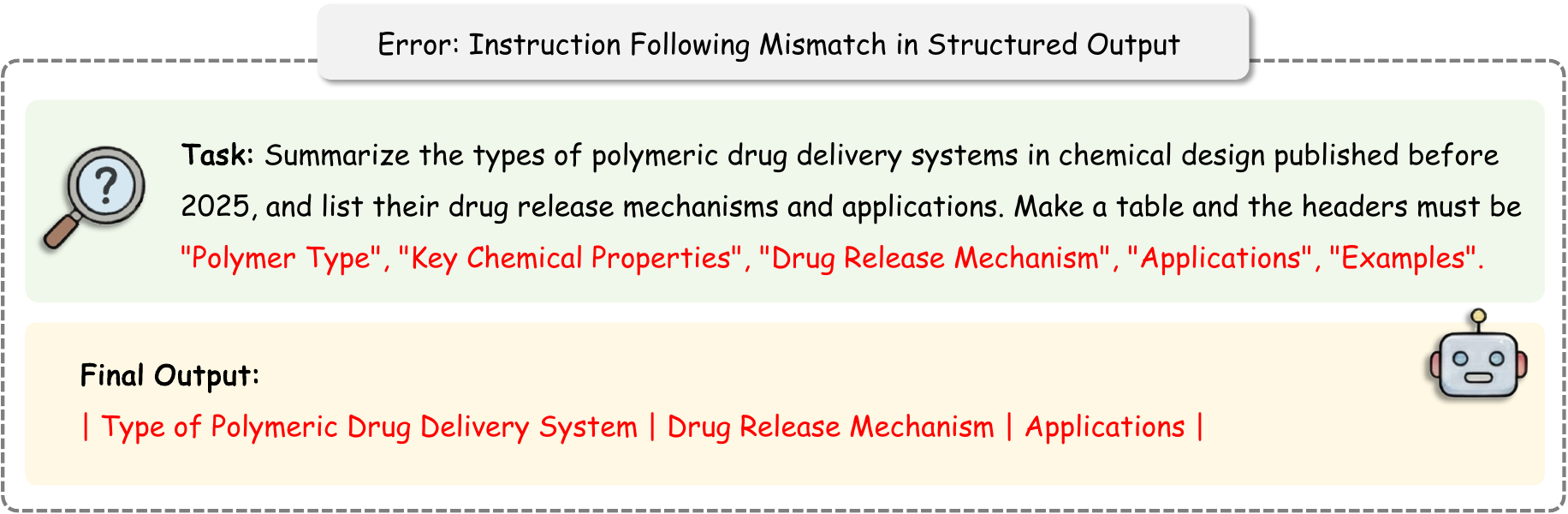}
    \caption{Structural Instruction Non-Compliance. The agent omits required columns ("Key Chemical Properties" and "Examples"), failing to adhere to the strict table schema provided.}
    \label{fig:e4}
\end{figure*}

\noindent\textbf{Case 4: Structural Instruction Non-Compliance} Figure \ref{fig:e4} presents a failure in following strict structural constraints during synthesis. The task explicitly required the generation of a Markdown table with five specific headers: "Polymer Type," "Key Chemical Properties," "Drug Release Mechanism," "Applications," and "Examples." While the agent successfully retrieved relevant content regarding polymeric drug delivery systems, the final output completely omitted the "Key Chemical Properties" and "Examples" columns, collapsing the information into a simplified three-column table. This exemplifies a schema alignment failure, where the model's internal training bias toward generating generic summaries overrides the explicit formatting constraints provided in the prompt. In automated scientific workflows where downstream parsers rely on fixed schemas, such structural instability renders the agent unusable.

\noindent\textbf{Case 5: Premature Hypothesis Commitment under Search.
} 
Figure~\ref{fig:e5} shows a failure case where search augmentation degrades performance by amplifying a hypothesis-first reasoning bias. Although the task explicitly specified a hard constraint on the target protein length (604 aa), the model prematurely inferred the pain-associated sodium channel Nav1.8 (SCN10A) based on superficial semantic cues. When search results later indicated that the canonical SCN10A protein is approximately 1956 aa, the model failed to revise its hypothesis and instead fabricated a non-existent “604 aa SCN10A construct” to satisfy the constraint. This behavior reflects post-hoc rationalization rather than evidence-driven hypothesis updating, where retrieved information is selectively distorted to justify an internally committed conclusion. Such failures highlight a key limitation of current search-augmented LLMs: external retrieval does not guarantee improved correctness when constraint enforcement and hypothesis revision mechanisms are absent.

\begin{figure*}[t]
    \centering
    \includegraphics[width=\linewidth]{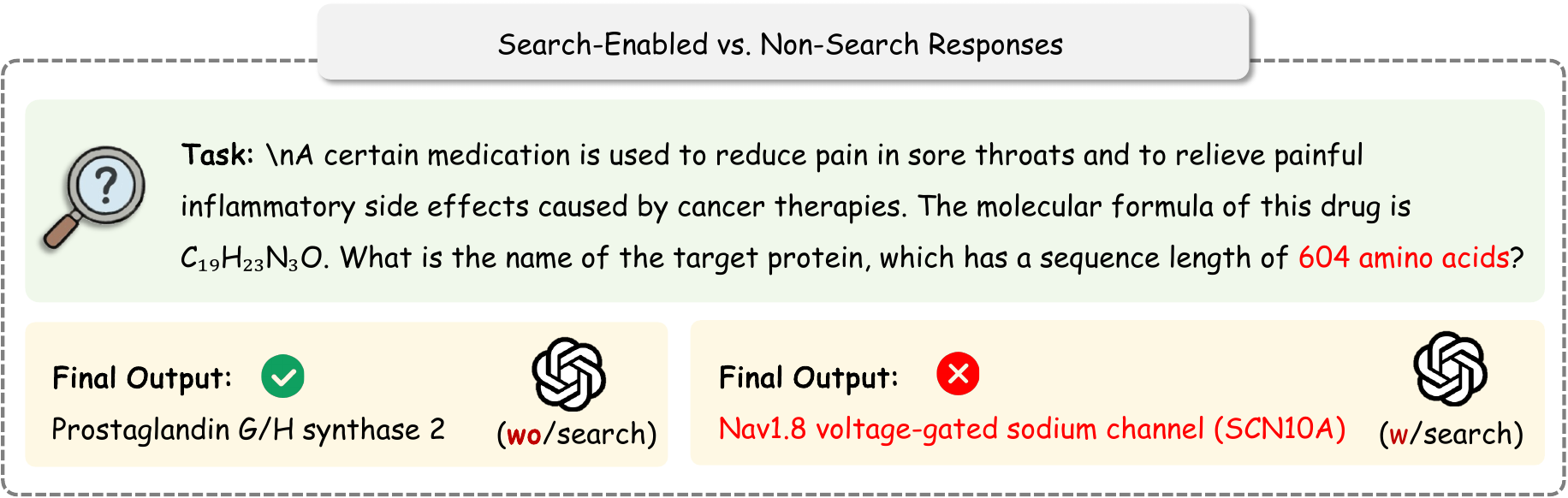}
    \caption{Comparison between search-enabled and non-search responses on a constraint-sensitive biomedical query. While the non-search model correctly identifies Prostaglandin G/H synthase 2 by adhering to the explicit sequence-length constraint (604 aa), the search-enabled model prematurely commits to a pain-related hypothesis (Nav1.8 / SCN10A) and fails to revise it after encountering contradictory evidence, resulting in an incorrect answer.}
    \label{fig:e5}
\end{figure*}

\begin{figure*}[t]
    \centering
    \includegraphics[width=\linewidth]{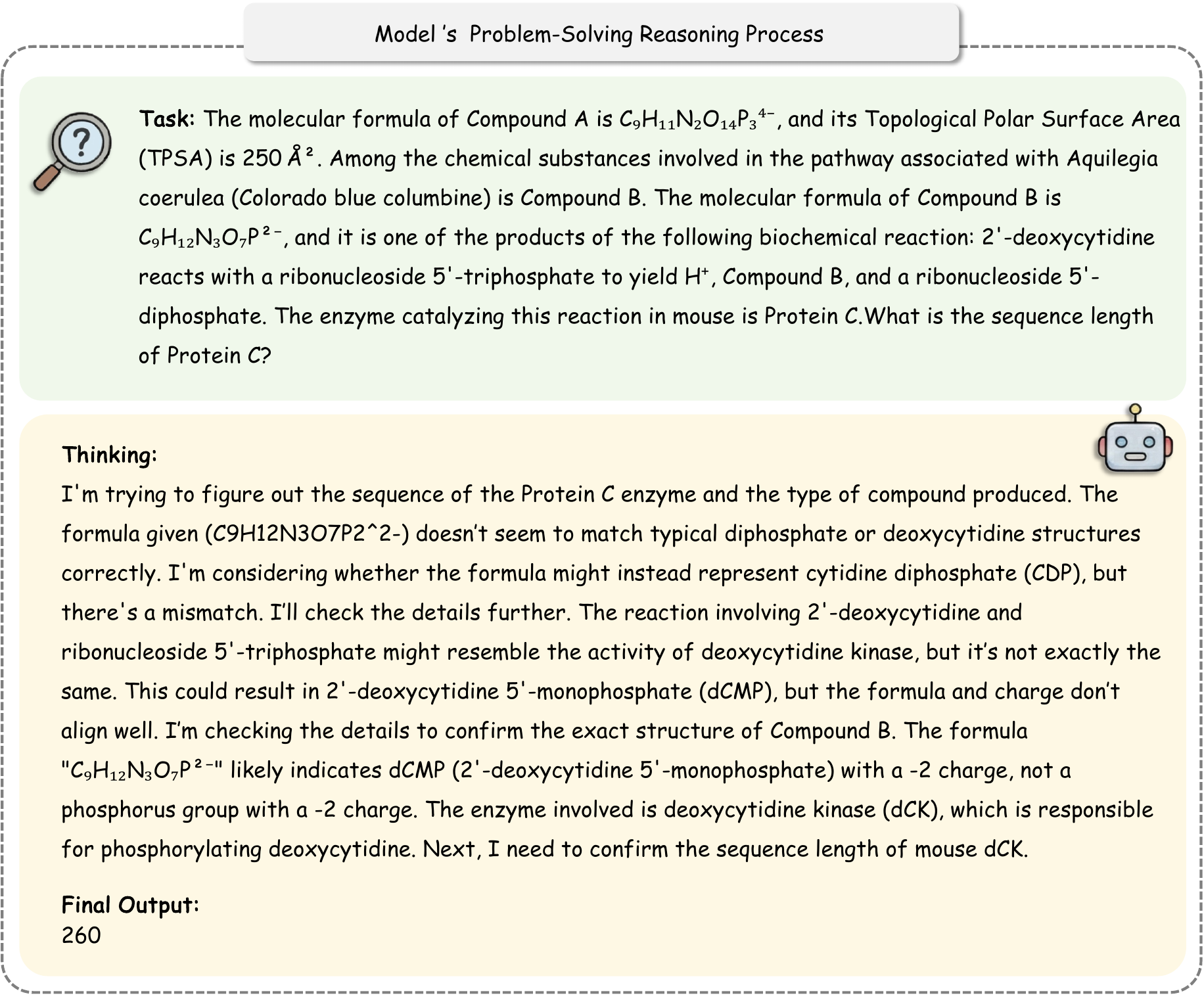} 
    \caption{Illustration of hypothesis-driven reasoning during model inference.
The model first proposes candidate interpretations for Compound B based on partial chemical cues, and subsequently validates these hypotheses against reaction mechanisms and known enzyme functions. The final answer is obtained after this verification stage, exemplifying a hypothesis-driven (guess-then-check) problem-solving process.}
    \label{fig:e6}
\end{figure*}

\subsection{Representative QA Examples of \textsc{SciExplore}}
\label{app:qa_examples}

% In this section, we present representative question--answer examples for each task type (T1--T4) in \textsc{SciExplore}.
In this section, we present representative question--answer examples for each task type (T1--T4) in \textsc{SciExplore}, illustrating the diverse reasoning and information-seeking behaviors required by the benchmark. These examples highlight how tasks progressively increase in complexity, ranging from structured database navigation and ambiguous literature retrieval to missing citation completion and cross-source structured knowledge synthesis. By examining concrete model inputs and outputs, we aim to provide an intuitive understanding of the challenges posed by each task type and the typical failure modes observed in current systems. Representative examples for T1, T2, T3, and T4 are shown in Figures~\ref{fig:example-t1},~\ref{fig:example-t2},~\ref{fig:example-t3},~\ref{fig:example-t4}, respectively.

\subsection{Prompt Templates}
We present the prompt templates used for each component in the \textsc{SciExplore} evaluation pipeline at the end of Appendix.

\newcounter{promptbox}
\renewcommand{\thepromptbox}{P\arabic{promptbox}}

\clearpage
\begin{figure*}[t]
\centering
\begin{examplebox}
\textbf{Question.}  
The molecular formula of Compound A is C$_9$H$_{11}$N$_2$O$_{14}$P$_3^{4-}$, and its Topological Polar Surface Area (TPSA) is 250~\AA$^2$.  
Among the chemical substances involved in the pathway associated with \textit{Aquilegia coerulea} (Colorado blue columbine) is Compound B.  
The molecular formula of Compound B is C$_9$H$_{12}$N$_3$O$_7$P$^{2-}$, and it is one of the products of the following biochemical reaction:  
2'-deoxycytidine reacts with a ribonucleoside 5'-triphosphate to yield H$^+$, Compound B, and a ribonucleoside 5'-diphosphate.  
The enzyme catalyzing this reaction in mouse is Protein C.  
What is the sequence length of Protein C?  
[The prediction error range is $\pm 0$]

\vspace{0.5em}
\textbf{Answer.}  
260
\end{examplebox}
\caption{\textbf{T1 Example Scientific Database Navigation}.}
\label{fig:example-t1}
\end{figure*}

\begin{figure*}[t]
\centering
\begin{examplebox}
\textbf{Question.}  
A research paper obtained various mixed crystals by adjusting the feed ratio.  
In the experiments of the paper, the surface growth method was used.  
The paper contains six figures in total, was published after 2018, and has at least one author with the surname Wang.  
What is the title of the paper?

\vspace{0.5em}
\textbf{Answer.}  
A solid-solution approach for controllable photomechanical crystalline materials
\end{examplebox}
\caption{\textbf{T2 Example Ambiguous Literature Retrieval}.}
\label{fig:example-t2}
\end{figure*}

\begin{figure*}[t]
\centering
\begin{examplebox}
\textbf{Question.}  
I would like to write an article on plastic pollution.  
The introduction section is already complete, but the sources supporting some key conclusions are still missing—please help me add them.

Plastic chemicals may also impede the transition to a circular economy, including technological solutions to plastic pollution[ref.1].  
For instance, increasing the reuse of plastic products can increase the release of chemicals[ref.2], and uncontrolled recycling can further perpetuate the spread of hazardous chemicals in products with high exposure potential, such as food packaging and toys.  
Chemicals in plastics can also hinder sorting and recycling and thereby impede the production of high-quality secondary materials[ref.3].  

Previous efforts to compile an overview of the chemical diversity of plastics[ref.4] have jointly identified more than 13,000 known plastic chemicals.

\textbf{Problem description:}  
All in-text citations appear as ``[ref.NUMBER]''.  
Please supply the complete bibliographic information for every citation, adhering to the specifications below.

\vspace{0.5em}
\textbf{Answer.}  

[ref.1: Wang, Z. \& Praetorius, A. Integrating a chemicals perspective into the global plastic treaty. Environ. 2022.]  

[ref.2: Mannoni, V. et al. Migration of formaldehyde and melamine from melaware and other amino resin tableware in real life service. \textit{Food Addit. Contam.} 2017.]  

[ref.3: Wiesinger, H., Wang, Z. \& Hellweg, S. Deep dive into plastic monomers, additives, and processing aids. 2021.]  

[ref.4: Aurisano, N., Weber, R. \& Fantke, P. Enabling a circular economy for chemicals in plastics. 2021.]
\end{examplebox}
\caption{\textbf{T3 Example Missing Citation Completion}.}
\label{fig:example-t3}
\end{figure*}

\begin{figure*}[t]
\centering
\begin{examplebox}
\textbf{Question.}

Find representative LLMs applied in bioinformatics in 2024, and list a table showing their Model name, Venue name, Model type, and Task type.

Here, the Model type includes three categories: Encoder-only, Decoder-only, Encoder-Decoder; if none of these apply, use - to indicate.
The Task type includes four categories: DNA, RNA, protein, scRNA.

Please organize the table in Markdown format.

\vspace{0.5em}
\textbf{Answer.}

\begingroup
\footnotesize
\setlength{\tabcolsep}{3pt}
\begin{verbatim}
| Model                  | Venue                           | Type            | Task    |
|------------------------|---------------------------------|-----------------|---------|
| RNA-MSM                | Nucleic Acids Research          | Encoder-only    | RNA     |
| GROVER                 | Nature Machine Intelligence     | Encoder-only    | DNA     |
| megaDNA                | Nature Communications           | Decoder-only    | DNA     |
| Nucleotide Transformer | Nature Methods                  | Encoder-only    | RNA     |
| RNAErnie               | Nature Machine Intelligence     | Encoder-only    | RNA     |
| RhoFold+               | Nature Methods                  | Encoder-only    | RNA     |
| GPTCelltype            | Nature Methods                  | Decoder-only    | scRNA   |
| Evo                    | Science                         | Decoder-only    | DNA     |
| scFoundation           | Nature Methods                  | Encoder-Decoder | scRNA   |
| scGPT                  | Nature Methods                  | Encoder-Decoder | scRNA   |
| AlphaFold3             | Nature                          | -               | Protein |
| ESM-DBP                | Nature Communications           | Encoder-only    | Protein |
| RoseTTAFold All-Atom   | Science                         | Encoder-Decoder | Protein |
| ProstT5                | NAR Genomics and Bioinformatics | Encoder-Decoder | Protein |
| EpiGcPT                | NAR Genomics and Bioinformatics | Encoder-only    | DNA     |
| RiNALMo                | arXiv                           | Encoder-only    | RNA     |
| EMBED                  | Bioinformatics Advances         | Encoder-Decoder | DNA     |
\end{verbatim}
\endgroup
\end{examplebox}
\caption{\textbf{T4 Example Cross-Source Structured Knowledge Synthesis
}.}
\label{fig:example-t4}
\end{figure*}

\onecolumn

\refstepcounter{promptbox}

\begin{tcolorbox}[
    breakable,             
    % enhanced,   
    % enhanced jigsaw,
    rounded corners,
    width=\textwidth,
    title={T1 and T2 JUDGE PROMPT},
    fonttitle=\normalsize\bfseries,
    coltitle=white,
    toptitle=2mm, bottomtitle=2mm, left=3mm, right=3mm, top=3mm, bottom=3mm,
]

\subsection*{Task Description}

As a grading expert, your task is to determine whether the candidate's final answer matches the provided standard answer. You must distinguish between exact matches, acceptable variations, and incorrect responses based on the guidelines below.

\subsection*{Evaluation Protocol}

Please follow these guidelines precisely:

\begin{enumerate}
    \item \textbf{Reference Standard:}
    \begin{itemize}
        \item The standard answer is definitive but allows for a certain margin of error.
        \item Do not regenerate answers; only compare with the given standard.
    \end{itemize}

    \item \textbf{Comparison Method:}
    \begin{itemize}
        \item Carefully analyze the question's requirements and the standard answer's structure:
        \begin{itemize}
            \item Determine whether the question expects \textbf{exact matching} or allows \textbf{partial matching}.
            \item Base this determination on the question's phrasing and the nature of the standard answer.
        \end{itemize}
        \item Compare \textbf{ONLY} the candidate's final answer (ignore reasoning/explanation errors).
        \item Disregard formatting or presentation style differences.
        \item Use discretion for unit omission (avoid automatic rejection).
        \item \textbf{Math:} Calculate step-by-step to check if formulas are equivalent.
        \item \textbf{Multiple Choice:} Compare only the final choice and option content.
    \end{itemize}

    \item \textbf{Multi-part Answers:}
    \begin{itemize}
        \item For questions requiring multiple responses (e.g., multi-select): All parts must match the standard answer exactly.
        \item Compare each sub-answer step by step. Partial matches are considered incorrect.
    \end{itemize}

    \item \textbf{Validity Check:}
    Reject answers that meet the following criteria (specify type in judgment):
    \begin{itemize}
        \item \textbf{Incomplete:} Cut off mid-sentence or lacks a complete response $\rightarrow$ Label as \textbf{INCOMPLETE}.
        \item \textbf{Repetitive:} Loops words or phrases $\rightarrow$ Label as \textbf{REPETITIVE}.
        \item \textbf{Refusal:} Explicit refusals (e.g., "I cannot answer...") $\rightarrow$ Label as \textbf{REFUSAL}.
    \end{itemize}
\end{enumerate}

\subsection*{Grading Scale}

\begin{itemize}
    \item $\boxed{A}$ \textbf{- CORRECT:}
    \begin{itemize}
        \item Answer matches standard exactly (including equivalent expressions).
        \item Within the allowable error range specified by the question.
        \item Semantically equivalent responses.
    \end{itemize}

    \item $\boxed{B}$ \textbf{- INCORRECT:}
    \begin{itemize}
        \item Outside the allowable error range specified by the question.
    \end{itemize}

    \item $\boxed{C}$ \textbf{- INCOMPLETE/REPETITIVE/REFUSAL:}
    \begin{itemize}
        \item Fails validity criteria (must specify subtype).
    \end{itemize}
\end{itemize}

\section*{Output Format}

Please strictly use the following format for your response:

\begin{lstlisting}[breaklines=true, basicstyle=\ttfamily]
Analysis step by step: [
Thoroughly evaluate the candidate's answer including:
(1) First check if the answer is INCOMPLETE, REPETITIVE, or a REFUSAL - if so, immediately classify as \boxed{C} with the corresponding type.
(2) Analyze the question's core requirements and the standard answer's structure (strict requirements vs. tolerant allowances).
(3) Perform a detailed comparison between the candidate's final answer and the standard answer (content equivalence, numerical precision, expression formats).
]
Final Judgment: \boxed{A/B/C} - <CORRECT/INCORRECT/INCOMPLETE/REPETITIVE/REFUSAL>
\end{lstlisting}

\subsection*{Task Input}

\texttt{<Original Question Begin>}\\
\texttt{\{question\}}\\
\texttt{<Original Question End>}

\vspace{0.2cm}

\texttt{<Standard Answer Begin>}\\
\texttt{\{gold\_answer\}}\\
\texttt{<Standard Answer End>}

\vspace{0.2cm}

\texttt{<Candidate's Answer Begin>}\\
\texttt{\{llm\_response\}}\\
\texttt{<Candidate's Answer End>}

\end{tcolorbox}

\begin{tcolorbox}[
    breakable,              % 允许跨页
    enhanced,
    rounded corners,
    width=\textwidth,
    title={T4 JUDGE PROMPT},
    fonttitle=\normalsize\bfseries,
    coltitle=white,
    toptitle=2mm, bottomtitle=2mm, left=3mm, right=3mm, top=3mm, bottom=3mm,
]

\subsection*{Role}

You are a strict evaluator for \textbf{single-row item-level recall} against a predicted table.

\subsection*{Inputs}

You will receive:

\begin{itemize}
    \item \textbf{Gold Row Table (ANSWER)}: Contains \textbf{a header + exactly one data row}.\\
    \texttt{\{ANSWER\_TABLE\}}
    
    \item \textbf{Predicted Table (PREDICTION)}: Contains a header + multiple rows (or may be empty).\\
    \texttt{\{PREDICTION\_TABLE\}}
\end{itemize}

\subsection*{Objective}

Determine whether PREDICTION contains the gold row (by primary key), then compute \textbf{item-level recall} for that single gold row.

\begin{itemize}
    \item The \textbf{first column} of ANSWER is the \textbf{primary key}.
    \item If the gold primary key is \textbf{not present} in any prediction row (after normalization), then \textbf{recall = 0.0}.
    \item If the primary key \textbf{is present}, compare the remaining cells (items) and compute:
\end{itemize}

Let:
\begin{itemize}
    \item \texttt{items\_total} = number of gold cells in the gold row \textbf{excluding} the primary key column, excluding empty/NA.
    \item \texttt{items\_recalled} = number of those cells that are matched by the predicted row with the same primary key (after column mapping + normalization).
\end{itemize}

Then:
\[ \texttt{recall} = \frac{\texttt{items\_recalled}}{\texttt{items\_total}} \]

Return \texttt{recall} as a float (e.g., 0.2) plus minimal supporting details.

\subsection*{Normalization Rules (apply BEFORE matching)}

\begin{enumerate}
    \item Ignore case, extra whitespace, and trivial punctuation.
    \item \textbf{Numeric tolerance}:
    \begin{itemize}
        \item Match if absolute diff $\le 10^{-6}$ OR relative diff $\le 10^{-3}$ (when $|\text{value}| > 10^{-6}$).
        \item Treat \texttt{0.25} as equal to \texttt{25\%} if clearly percentage-formatted.
    \end{itemize}
    \item \textbf{Units}: convert when unambiguous (e.g., g $\leftrightarrow$ mg). If units conflict, treat as mismatch.
    \item \textbf{Aliases}: allow common aliases if clearly the same entity.
    \item \textbf{List-in-cell}:
    \begin{itemize}
        \item If a cell contains a list separated by commas/semicolons, treat as a set.
        \item Count as recalled only if predicted set contains all gold elements (subset match).
    \end{itemize}
    \item \textbf{Be conservative}: if uncertain, mark mismatch.
\end{enumerate}

\subsection*{Column Mapping}

\begin{itemize}
    \item Use ANSWER headers as the canonical schema.
    \item Map each gold column to the best matching prediction column by:
    \begin{enumerate}
        \item exact header match after normalization, else
        \item semantic similarity of header names.
    \end{enumerate}
    \item If a gold column cannot be mapped, that item is \textbf{not recalled}.
\end{itemize}

\subsection*{Matching Procedure (must follow)}

\begin{enumerate}
    \item Parse ANSWER: extract headers and the single gold row.
    \item Identify $\texttt{pk\_col} = \text{first header}$, $\texttt{gold\_pk} = \text{normalized}(\text{value in pk\_col})$.
    \item Parse PREDICTION rows and build an index from normalized pk values to candidate rows.
    \item If \texttt{gold\_pk} not found in prediction index:
    \begin{itemize}
        \item Set \texttt{items\_total} as defined above.
        \item Set $\texttt{items\_recalled} = 0$.
        \item Set $\texttt{recall} = 0.0$.
    \end{itemize}
    \item If \texttt{gold\_pk} found:
    \begin{itemize}
        \item If multiple predicted rows share the same pk, choose the row that maximizes \texttt{items\_recalled}.
        \item For each eligible gold item cell:
        \begin{itemize}
            \item Compare gold cell vs predicted mapped cell after normalization.
            \item Count match as recalled; otherwise mismatch.
        \end{itemize}
    \end{itemize}
\end{enumerate}

\section*{Output Format (STRICT JSON ONLY)}

Return ONLY one JSON object:

\begin{lstlisting}[breaklines=true, basicstyle=\ttfamily, frame=none]
{
  "recall": float,
  "items_recalled": int,
  "items_total": int,
  "primary_key_column": "string",
  "gold_primary_key": "string (normalized)",
  "matched_prediction_row_index": int | null,
  "column_mapping": { "gold_col": "pred_col_or_null", ... },
  "mismatches": [
    {
      "column": "gold column name",
      "gold_value": "normalized gold cell",
      "pred_value": "normalized predicted cell or null",
      "reason": "short reason"
    }
  ]
}
\end{lstlisting}

\subsection*{Important Constraints}

\begin{itemize}
    \item Do not output any text outside the JSON.
    \item Do not hallucinate missing values.
    \item If $\texttt{items\_total} == 0$, set $\texttt{recall} = 0.0$.
\end{itemize}

\end{tcolorbox}

\end{document}